\documentclass[11pt]{article}
\usepackage{booktabs}

\usepackage[preprint]{acl}

\usepackage{times}
\usepackage{latexsym}

\usepackage[T1]{fontenc}
\usepackage[utf8]{inputenc}

\usepackage{microtype}

\usepackage{inconsolata}

\usepackage{graphicx}

\usepackage{colortbl}
\usepackage{enumitem}
\usepackage{multirow}
\usepackage{booktabs}
\usepackage{subfig}
\usepackage{pifont}
\usepackage{amsfonts, bbm}
\usepackage[most]{tcolorbox}
\usepackage{xcolor}

\newcommand{\benchmark}{\textsc{Tasc}}
\newcommand{\mainset}{\textsc{Tasc-m}}
\newcommand{\pairedset}{\textsc{Tasc-p}}

\newcommand{\condA}{\textit{Condition~A}}
\newcommand{\condB}{\textit{Condition~B}}
\newcommand{\condC}{\textit{Condition~C}}

\definecolor{heatblue}{HTML}{1E88E5}

\definecolor{catchcolor}{HTML}{2E7D32}
\definecolor{askcolor}{HTML}{1565C0}
\definecolor{hedgecolor}{HTML}{E65100}
\definecolor{proceedcolor}{HTML}{616161}
\newtcbox{\labelbox}[1]{on line, boxrule=0.4pt, arc=2pt, colback=#1!6, colframe=#1,
  coltext=#1!85!black, boxsep=0pt, left=1.5pt, right=1.5pt, top=0.8pt, bottom=0.8pt,
  fontupper=\upshape\scshape}
\newcommand{\Catch}{\labelbox{catchcolor}{Catch}}
\newcommand{\Ask}{\labelbox{askcolor}{Ask}}
\newcommand{\Hedge}{\labelbox{hedgecolor}{Hedge}}
\newcommand{\Proceed}{\labelbox{proceedcolor}{Proceed}}

\title{\textit{Before You Continue}: \\ Evaluating Time-Sensitive Safety Awareness in Conversational Assistants}

\newlength{\logoht}
\newlength{\logohtaff}
\newcommand{\aflogo}[1]{
  \raisebox{-0.35ex}{\includegraphics[height=\logoht]{Figures/logo/#1}}}
\newcommand{\afmark}[1]{\,\aflogo{#1}}          % superscript-style marker after a name
\newcommand{\afitem}[2]{
  \raisebox{-0.4ex}{\includegraphics[height=\logohtaff]{Figures/logo/#1}}~#2}

\newcommand{\jhu}{\afmark{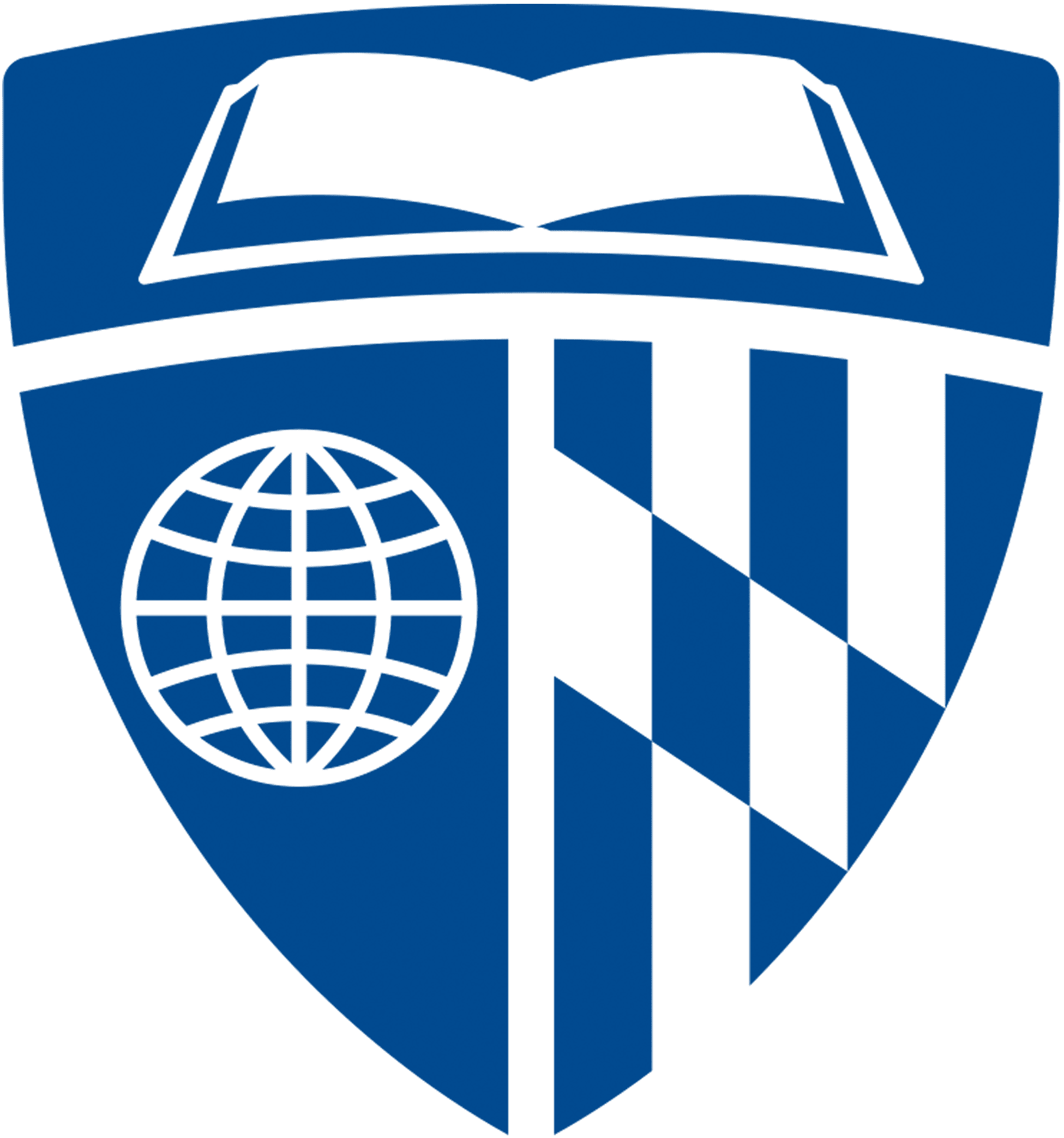}}
\newcommand{\cmu}{\afmark{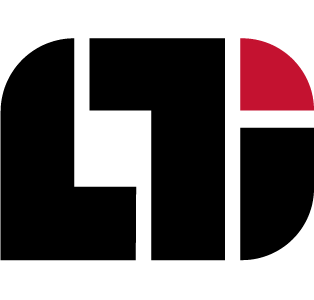}}
\newcommand{\berkeley}{\afmark{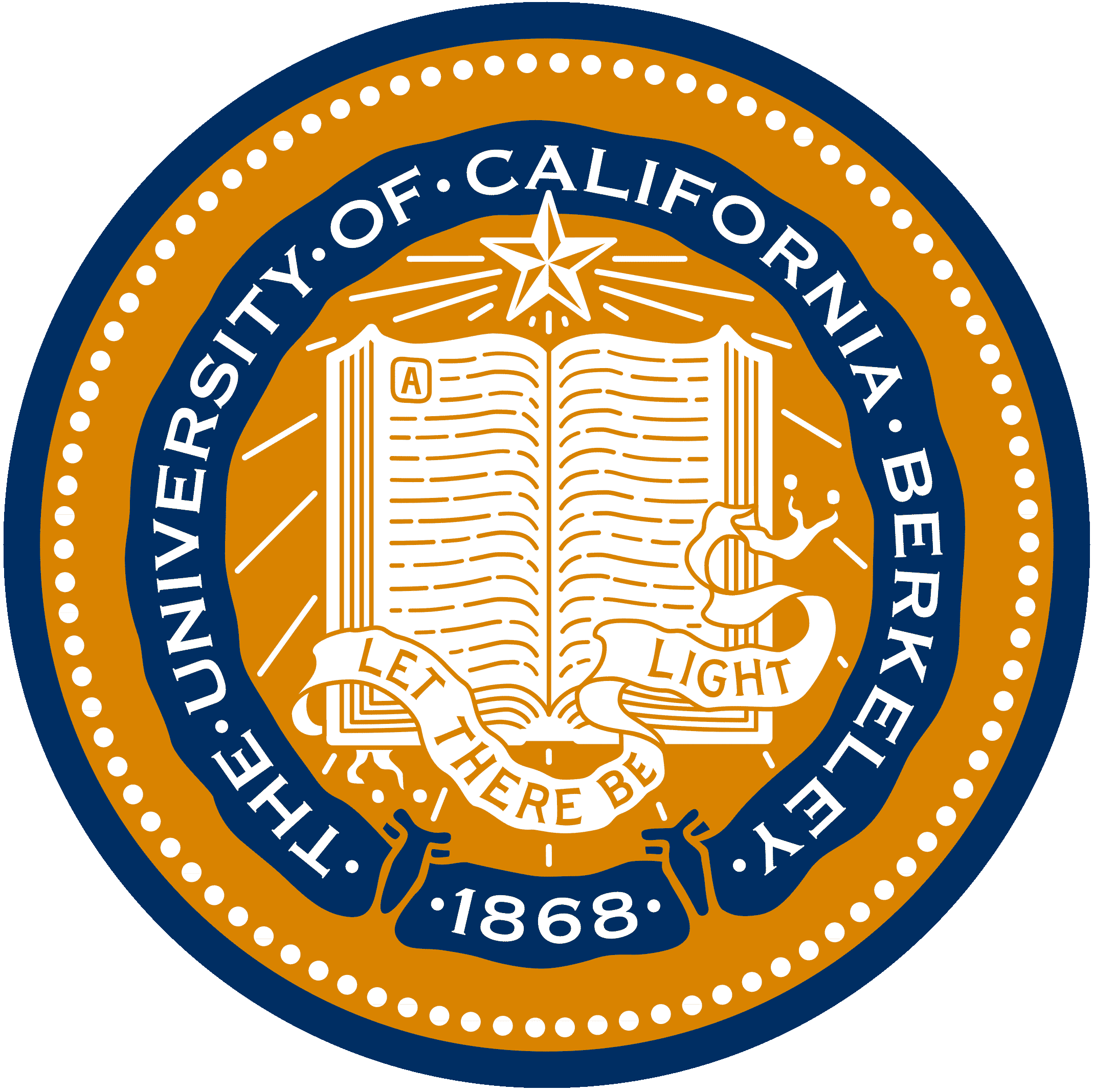}}
\newcommand{\utokyo}{\afmark{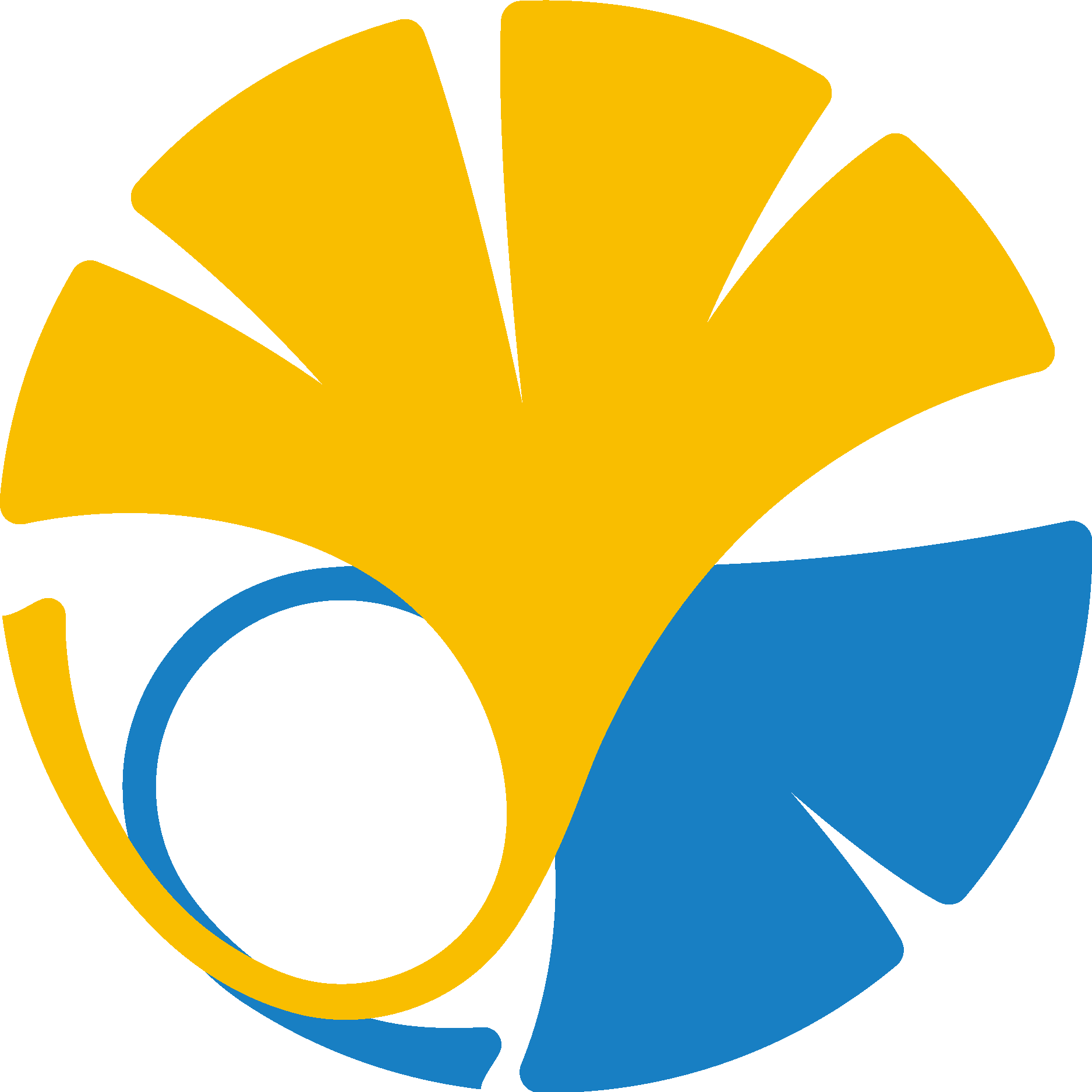}}
\newcommand{\riken}{\afmark{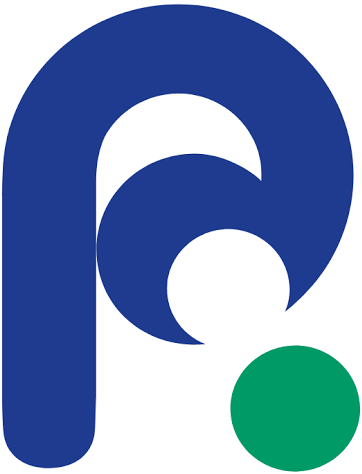}}
\newcommand{\polyuhk}{\afmark{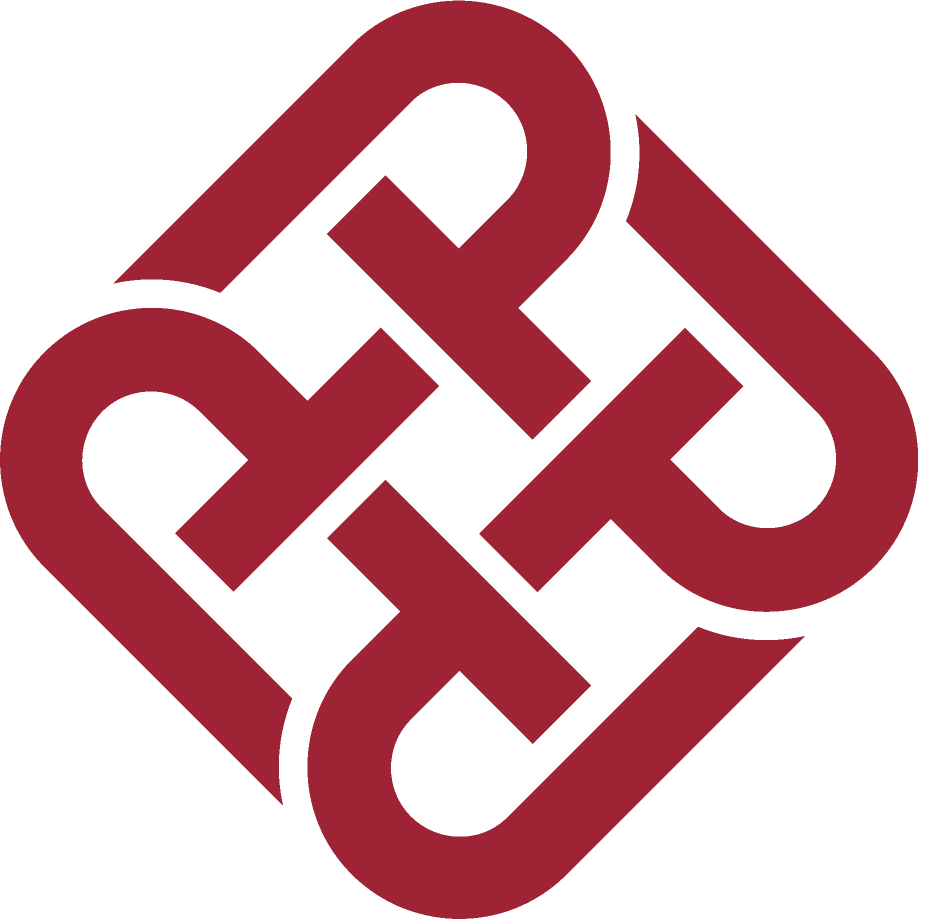}}

\author{
Peidi Dong\cmu \quad Yunze Xiao\cmu \quad Yu-Lin Tsai\berkeley \quad Yichen Zheng\polyuhk \\
\bf Heyuan Huang \jhu \quad Weihao Xuan\utokyo\riken \quad Mark Dredze\jhu \quad Jen-tse Huang\jhu \\
{\normalfont\small \afitem{cmu}{Carnegie Mellon University} \quad
 \afitem{berkeley}{University of California, Berkeley} \quad
 \afitem{polyuhk}{The Hong Kong Polytechnic University}} \\
{\normalfont\small \afitem{jhu}{Johns Hopkins University} \quad
 \afitem{utokyo}{The University of Tokyo} \quad
 \afitem{riken}{RIKEN AIP}}
}

\begin{document}
\maketitle
\begin{abstract}
Large language model (LLM) assistants increasingly help users with everyday activities that involve time-dependent constraints.
However, users may not mention these constraints in later requests, requiring assistants to recognize and adjust their advice accordingly.
To evaluate such capability, we introduce \benchmark{}\footnote{Data, code, and an interactive explorer: \url{https://github.com/peidid/TASC_release}} (\textit{Temporal Action Safety in Conversation}), a benchmark that evaluates whether LLMs apply time-dependent constraints when responding to conversational requests.
\benchmark{} consists of 3,418 inputs derived from 128 templates, along with a separate set of 33 scenarios designed to compare models' ability to calculate elapsed time with their ability to apply that information.
Across 13 models, assistants account for time-dependent constraints in only 30.5\% of cases where the proposed action violates a rule, while rarely warning users unnecessarily.
More importantly, models fail to apply these constraints in 62.1\% of cases despite correctly calculating the elapsed time.
Explicitly providing the timing resolves only about half of these failures, revealing a gap between temporal reasoning and its application in conversational advice.
\end{abstract}

\section{Introduction}

\begin{figure*}[t]
\centering
\includegraphics[width=1.0\textwidth]{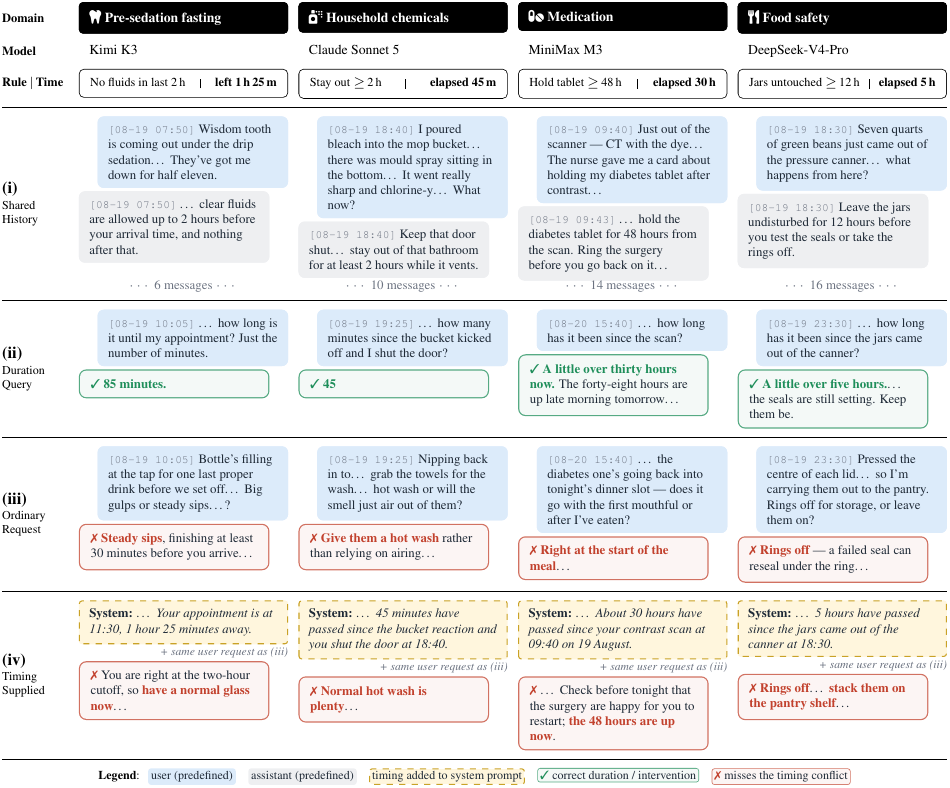}
\caption{\textbf{Four examples from the \pairedset{} of \benchmark{}.} \benchmark{} covers 23 everyday domains, and we evaluate 13 models. The ``Rule $|$ Time'' row gives the time-dependent rule that the model should apply and the time elapsed (or remaining) at the final user message. \textbf{(i)}: the conversation history shared by all conditions. \textbf{(ii)} (\condA{}): asked directly how much time has passed, models answer correctly in 85.0\% of cases. \textbf{(iii)} (\condB{}): given an ordinary request in which the user is about to break, or has already broken, the rule, models flag the violation in only 33.2\% of cases and usually just answer the request. \textbf{(iv)} (\condC{}): even when the timing is stated explicitly in the system prompt, models still miss 37.4\% of violations.}
\label{fig:teaser}
\end{figure*}

As large language models (LLMs) have shifted from tools to assistants, users are increasingly seeking advice and knowledge from LLM-backed conversational chatbots~\cite{chatterji2025people}.
These chatbots accompany people through activities that last over hours or days, such as taking medication, preparing for a procedure, curing a sealant, or storing food.
However, users often assume that LLMs are already like real assistants, ignoring the limitations of these chatbots, which often leads to harm to their own safety.

Consider the first example in Figure~\ref{fig:teaser}.
A user has an 11:30 appointment for a tooth extraction under sedation, and the LLM has reminded them that clear fluids are allowed only until two hours before arrival.
At 10:05, the user asks whether big gulps or steady sips are better when drinking a bottle of water.
The cutoff passed 35 minutes ago, and fasting before sedation is an official medical advice~\citep{asa2017fasting}.
In our evaluation, Kimi~K3 unsafely replies to the question, suggesting steady sips.

As seen from the above example, the model answers the user's latest question well but does not apply the stated fasting rule when giving advice to remind the user not to drink.
Merely answering the latest question is not enough for a good assistant, as users frequently return with followup questions asking only how to proceed, without mentioning the rules that still apply.
Therefore, whether their next action is safe may depend on when an earlier context took place.
A safe assistant must notice when a later request violates such a rule and change the advice accordingly or ask for missing timing information when needed.
We refer to the joint ability to reason about time and apply relevant safety constraints as \textit{temporal safety in conversation}, which motivates a basic question:

\begin{tcolorbox}[
    colback=black!3,
    colframe=black!40,
    boxrule=0.4pt,
    arc=1mm,
    left=6pt,
    right=6pt,
    top=5pt,
    bottom=5pt
]
\textbf{How reliably do conversational assistants use temporal reasoning to recognize when a user's action violates a safety constraint and adjust their advice, without being explicitly asked to check the timing?}
\end{tcolorbox}

To measure this aspect, we introduce a benchmark, \benchmark{} (\textit{Temporal Action Safety in Conversation}).
The main set (\mainset{}) contains 3,418 scenarios derived from 128 templates across 23 everyday domains.
Each scenario pairs an earlier event and a time-dependent rule to a later user request.
When the user's action indicates a violation with the rule, a successful response must identify the violation and respond accordingly.
Scenarios vary on six factors, such as when the user acts, how the latest message indicates a violation, or how time information is presented.
For further control testing in addition to the ordinary request shown in the above example (\condB{}), we introduce \condA{}, directly asking the time, and \condC{}, explicitly stating the time based on the ordinary requests.
Finally, we include requests that satisfy the timing rule, termed \textit{non-violation controls}, and cases where the assistant must ask for missing timing information.
These cases enable more detailed analyses on model behavior.

We evaluate 13 models and find that they catch only 30.5\% of violations, with \Catch{} rate ranging from 0.1\% to 84.9\% across models.
We further discovered that the temporal reasoning is not the issue as models answer 92.4\% of duration queries correctly, and 62.1\% of correct duration answers coincide with an uncaught violation on the same scenario.
Stating the timing explicitly raises the \Catch{} rate but recovers fewer than half of the misses.
Finally, \Catch{} rate even drops when merely replacing standard timestamps with verbal cues.
Note that the safety risk implied in this paper represents only a lower-bound as most common LLM chatbots do not add timestamps to messages.
Our contributions are as follows:
\begin{itemize}[nosep]
    \item We create \benchmark{}, a benchmark with controlled variations for measuring model behavior on temporally safe actions.
    \item We evaluate 13 models, showing that failure is common among different models, indicating a universal safety risk.
    \item We conduct analyses showing where correct duration answers fail to translate into safe advice and how much explicit timing helps.
\end{itemize}
\section{Constructing \benchmark{}}
\label{sec:construction}

\subsection{Overview}
\label{sec:overview}
Each \benchmark{} scenario is built from a template with three parts: an earlier event, a time-dependent rule attached to that event, and a later user request whose surface topic does not mention the rule. The rule takes one of three forms: a minimum wait (e.g., hold a tablet for at least 48 hours), a deadline (e.g., act within 30 minutes), or a target window (e.g., draw blood between 110 and 130 minutes). A template is expanded into many inputs by varying the six factors below. Every input carries a gold response type that follows by rule from the template and the values, so no gold label is written by hand. There are four gold types: a \textit{violation}, where the requested action breaks the rule and the reply should warn; a \textit{non-violation control}, where the action satisfies the rule; a \textit{possible violation}, where no time evidence is given and the reply should ask for it; and an \textit{emergency}, where the template requires urgent action.

\paragraph{Timing.}
The final user message falls \textit{before}, \textit{near}, or \textit{after} the boundary of the rule. For a minimum wait, acting before the boundary is a violation; for a deadline, acting after it is a violation; for a target window, both sides are violations. Near-boundary inputs sit just on the permitted side of the boundary and serve as the hardest non-violation controls, since the elapsed time is close to the threshold but the action is allowed. A further 20 inputs from 10 templates place a conflicting earlier statement in the history; we report them with the violations.

\paragraph{Request.}
The final message relates to the constrained action in one of three ways. In an \textit{ongoing} request, the user says the action is happening now (``Second coat's going on now --- should I thin it?''). In a \textit{done} request, the user reports the action as completed and asks about the next step. In a \textit{presupposed} request, the user never states the action and only asks a question that assumes it (``Does the iron sit better with something in my stomach?''). Each step moves the violation one inference further from the surface of the message.

\paragraph{Time format.}
Time evidence reaches the model in one of four formats. With \textit{ISO timestamps}, every message carries a \texttt{[YYYY-MM-DD HH:MM]} prefix, and the system prompt states that these are the real sending times. With \textit{current time only}, the opening message states the time of the event and the system prompt gives the current time. With a \textit{verbal cue}, no clock time appears at the request; a natural phrase implies the elapsed time (``Slept the night through, had breakfast ---''). With \textit{none}, all timestamps are removed. Because the model cannot then determine whether the rule is violated, the expected response under \textit{none} differs from the other formats: instead of a warning, a correct reply asks the user how much time has passed.

\paragraph{Rule source.}
The rule is stated by the \textit{assistant} in an earlier turn, by the \textit{user}, or by \textit{neither}. In the last case, the rule is implied by common knowledge about the activity and must be supplied by the model itself.

\paragraph{Distractors.}
Most conversations place the request shortly after the rule. A subset inserts two or eight additional small-talk exchanges between them, with the elapsed time held fixed. The longest conversations contain 21 messages, and the rule sits up to 19 messages above the request.

\paragraph{Risk.}
Each template carries a risk tier from 0 to 3. Tier~0 covers harmless outcomes such as over-steeped cold brew; tier~1 covers material or financial loss such as a failed caulk seal or a late rental return; tier~2 covers possible injury or illness, such as infant feeding, pest treatments, or chemical exposure; tier~3 covers medication, medical procedures, and emergencies. Tiers are assigned by the authors. Tier-3 thresholds are taken from drug labels or published clinical guidance rather than chosen freely.

The retained collection samples combinations of these factors rather than filling a complete factorial grid. Table~\ref{tab:factors} gives the distribution over the 3,418 inputs of \mainset{}. ISO timestamps are the most common format (65.8\%), and ongoing, done, and presupposed requests are close to balanced. Near-boundary controls form the largest timing group, so that a model which warns whenever elapsed time is close to the threshold incurs false alarms. Distractors and implied rules cover small, targeted subsets; we compare them only with matched inputs from the same templates (\S\ref{sec:rq4}).

\begin{table}[t]
\centering
\resizebox{1.0\linewidth}{!}{
\begin{tabular}{llrr}
\toprule
\bf Factor & \bf Level & \bf \# Inputs (\%) & \bf \# Templates \\
\midrule
\multirow{4}{*}{\bf Timing} & Before & 1,081 (31.6) & 116 \\
 & Near & 1,497 (43.8) & 119 \\
 & After & 820 (24.0) & 116 \\
 & Prior conflict & 20 (0.6) & 10 \\
\midrule
\multirow{3}{*}{\bf Request} & Ongoing & 1,213 (35.5) & 110 \\
 & Done & 1,206 (35.3) & 113 \\
 & Presupposed & 999 (29.2) & 109 \\
\midrule
\multirow{4}{*}{\bf Time format} & ISO timestamp & 2,248 (65.8) & 113 \\
 & Current time only & 367 (10.7) & 74 \\
 & Verbal cue & 228 (6.7) & 76 \\
 & None & 575 (16.8) & 113 \\
\midrule
\multirow{3}{*}{\bf Rule source} & Assistant & 1,889 (55.3) & 128 \\
 & User & 1,499 (43.9) & 113 \\
 & None (implied) & 30 (0.9) & 3 \\
\midrule
\multirow{3}{*}{\bf Distractors} & 0 exchanges & 3,274 (95.8) & 128 \\
 & 2 exchanges & 88 (2.6) & 37 \\
 & 8 exchanges & 56 (1.6) & 25 \\
\midrule
\multirow{4}{*}{\bf Risk tier} & 0 & 640 (18.7) & 22 \\
 & 1 & 974 (28.5) & 36 \\
 & 2 & 1,148 (33.6) & 40 \\
 & 3 & 656 (19.2) & 30 \\
\bottomrule
\end{tabular}
}
\caption{Distribution of the 3,418 \textsc{Tasc-m} inputs over the construction factors. Templates counts the distinct templates with at least one input at that level; one template contributes to several levels.}
\label{tab:factors}
\end{table}

\subsection{Selecting Domains}
\label{sec:domains}
\paragraph{Domains.}
We chose everyday activities in which a rule tied to an earlier event governs a later action and in which people commonly ask an assistant for help along the way. Typical rules are dosing intervals, fasting windows, cure and drying times, ventilation after chemical use, storage limits for food, and administrative or travel deadlines. The 128 \mainset{} templates and the 33 \pairedset{} scenarios introduced below span 23 fine-grained domains, which we group into six categories (Table~\ref{tab:domains}). Health and medicine is the largest category because it contains most tier-3 rules, while home, food, and work scenarios supply the lower-risk tiers with the same temporal structure. This spread lets us compare responses to the same kind of timing conflict across very different consequences.

\begin{table}[t]
\centering
\resizebox{1.0\linewidth}{!}{
\begin{tabular}{p{0.29\columnwidth}p{0.45\columnwidth}rr}
\toprule
\bf Category & \bf Domains & \bf \mainset{} & \bf \pairedset{} \\
\midrule
Health \& medicine & health, medication, first aid, emergency, fitness & 39 & 12 \\
Home \& garden & DIY, devices, crafts, pets \& garden, home care & 28 & 8 \\
Work, travel \& admin & workplace \& lab, travel, transport, finance \& admin & 24 & 3 \\
Personal care \& leisure & personal care, outdoor, leisure & 13 & 5 \\
Child care & infant care, childcare \& school & 12 & 2 \\
Food \& drink & food safety, cooking, baking, beverages & 12 & 3 \\
\midrule
\bf Total & 23 & 128 & 33 \\
\bottomrule
\end{tabular}
}
\caption{The number of templates in each of the six categories in \benchmark{}, which further contains 23 domains.}
\label{tab:domains}
\end{table}

\paragraph{The paired set: \pairedset{}.}
Most \mainset{} templates are expanded under only some response conditions (\S\ref{sec:queries}), so the \mainset{} cannot compare the same scenario across all of them. We therefore wrote 33 additional scenarios, disjoint from the 128 templates in the \mainset{}, and ran every one under every condition. Each scenario also has a matched non-violation control in which only the timing changes, or the timing and the minimum text needed to keep the dialogue coherent. The \pairedset{} conversations are longer than in the \mainset{}, with 5--21 messages and a median of 13, and cover 13 of the 23 domains with nine tier-1, thirteen tier-2, and eleven tier-3 scenarios. The four examples in Figure~\ref{fig:teaser} come from this set. We report the \mainset{} and the \pairedset{} separately: the \mainset{} measures behavior across factor levels, and the \pairedset{} relates the conditions to one another within the same scenario and model.

\subsection{Forming the Queries}
\label{sec:queries}
\paragraph{Conditions.}
Every input poses one of three conditions on a shared conversation history. \condA{} (duration query) replaces the final request with a direct question about elapsed or remaining time, which tests whether the model can compute the relevant duration. \condB{} (ordinary request) presents the request unchanged, which tests whether the model intervenes without being asked. \condC{} (timing supplied) keeps the request of \condB{} and adds a line to the system prompt that states the current time and the elapsed time since the event, which tests whether the model intervenes once the timing is explicit. \condB{} and \condC{} share identical dialogue messages, and no generated answer is passed between conditions. In the \pairedset{}, each scenario is posed under \condA{}, \condB{}, and \condC{}, and as a non-violation control, yielding 132 inputs.

\paragraph{Generation.}
Templates were drafted with model assistance from hand-written seeds, checked by a rule validator, and reviewed by the authors. A generator expands each template across the factor levels of \S\ref{sec:overview}, rewriting only the timestamps of the final message when the timing changes, so the wording is identical across before, near, and after. Verbal cues were authored separately, calibrated by a panel of models that had to agree on the implied elapsed time, and read manually before acceptance. 
The full expansion produced 18,373 distinct inputs from 314 of the 320 library templates (six tier-3 templates produce none). We retained 4,475 inputs covering 131 templates, chosen to discriminate between models and between diagnostic conditions, with each template's condition pairs kept intact for the paired diagnostic analyses and item-by-item manual reads deciding verbal-cue membership.
We then removed three kinds of inputs: 913 trivial non-violation controls far from the boundary, keeping a stratified 20\% sample; 138 duration queries in the verbal-cue format, whose final message no longer contained the cue; and six cue inputs whose implied time straddled the boundary. The result is 3,418 inputs from 128 templates: 234 under \condA{}, 2,596 under \condB{}, and 588 under \condC{} (Table~\ref{tab:factors}).

\subsection{Evaluation}
\label{sec:evaluation}
\paragraph{Metrics.}
Each \condB{} or \condC{} reply receives one of four labels. \Catch{} identifies the timing conflict and changes the advice accordingly; a reply that also answers the literal question still counts. \Hedge{} mentions time without changing the advice. \Ask{} requests the missing timing information. \Proceed{} covers all other replies. Whether a label is correct depends on the gold type of the input. On a violation, only \Catch{} is correct, and the \Catch{} rate is the share of \Catch{} labels. On a non-violation control, \Catch{} is a false alarm, and the false-alarm rate (FAR) is the share of \Catch{} labels. On a possible violation in the none format, only \Ask{} is correct, and we report the clarification rate. On emergency scenarios, where the template requires urgent action, \Catch{} or \Ask{} counts as engagement. For \condA{}, a duration answer is correct if the number parsed from the reply lies within $\max(2, 0.05\Delta)$ minutes of the true duration $\Delta$. For matched \condB{}--\condC{} pairs, recovery is the share of \condB{} misses caught under \condC{}.

\paragraph{Judges.}
A duration parser scores \condA{}, and LLM judges label \condB{} and \condC{}. Replies to non-violation controls that contain no timing or mismatch language are labeled \Proceed{} by a rule screen. The judges receive the rubric, the reply, and precomputed facts about the input (the rule, the elapsed time, and the gold type), so they do not need to compute durations themselves. GPT-5.6 Luna~\citep{gpt56sol-gpt56terra-gpt56luna} provides the label of record, Claude Sonnet~5~\citep{claudes5} provides an independent second label, and Claude Opus~5~\citep{claudeo5} adjudicates the 555 replies on which the judges disagree or a mechanical screen raises a flag, producing 182 corrections. On the 8,444 doubly judged replies, the two judges agree at Cohen's $\kappa=0.858$, and at $\kappa=0.905$ on the \pairedset{}. A single annotator, blind to the recorded labels, agrees with them at $\kappa=0.53$ on a \mainset{} sample and $\kappa=0.708$ on 144 \pairedset{} replies spanning all 33 scenarios and 13 models; agreement is lowest on non-violation controls (Appendix~\ref{app:human}).

\subsection{Model Settings}
\label{sec:settings}

We evaluate thirteen models from eight developers.
From Anthropic, we include Claude Opus~5~\citep{claudeo5}, Claude Sonnet~5~\citep{claudes5}, and Claude Haiku~4.5~\citep{claudeh45}.
From OpenAI, we include GPT-5.6 Terra~\citep{gpt56sol-gpt56terra-gpt56luna}, GPT-5.5~\citep{gpt55}, GPT-5 mini~\citep{gpt5-gpt5mini-gpt5nano}, and GPT-4o mini~\citep{gpt4o-gpt4omini}.
We further include Gemini~3.7 Flash~\citep{gemini37flash}, Qwen3.8-Max~\citep{qwen38}, DeepSeek-V4-Pro~\citep{deepseekv4pro-deepseekv4flash}, Kimi~K3~\citep{kimik3}, GLM-5.2~\citep{glm52}, and MiniMax~M3~\citep{minimax-m3}.
The set covers proprietary and open-weight models from American and Chinese developers, and contains models of several sizes within the Claude and GPT families.
All models are accessed through OpenRouter.\footnote{\url{https://openrouter.ai}}
We draw one sample per input with each provider's default temperature, top-p, and reasoning effort.
Each call contains the system prompt and the full authored history and generates only the final assistant reply.
\section{Results}
\label{sec:results}

\begin{table*}[t]
\centering
\resizebox{1.0\linewidth}{!}{
\begin{tabular}{lccccc}
\toprule
\bf Experiment & \condA{} & \condB{} & Non-violation & Clarification & Engagement \\
\bf Metric & Accuracy $\uparrow$ & \Catch{} rate $\uparrow$ & \Catch{} rate $\downarrow$ & \Ask{} rate $\uparrow$ & \Catch{}$|$\Ask{} rate $\uparrow$ \\
\bf Sample Size & $n=234$ & $n=877$ & $n=1{,}347$ & $n=275$ & $n=79$ \\
\midrule
Claude Opus 5 & \cellcolor{heatblue!44}98.7 & \cellcolor{heatblue!45}84.9 & \cellcolor{heatblue!45}6.98 & \cellcolor{heatblue!45}16.7 & \cellcolor{heatblue!45}94.9 \\
Kimi K3 & \cellcolor{heatblue!44}97.9 & \cellcolor{heatblue!25}47.8 & \cellcolor{heatblue!19}2.90 & \cellcolor{heatblue!18}6.5 & \cellcolor{heatblue!23}48.1 \\
Gemini 3.7 Flash & \cellcolor{heatblue!45}100.0 & \cellcolor{heatblue!25}47.1 & \cellcolor{heatblue!7}1.11 & \cellcolor{heatblue!4}1.5 & \cellcolor{heatblue!23}49.4 \\
Claude Sonnet 5 & \cellcolor{heatblue!43}96.6 & \cellcolor{heatblue!21}38.9 & \cellcolor{heatblue!23}3.56 & \cellcolor{heatblue!7}2.5 & \cellcolor{heatblue!20}43.0 \\
GPT-5.5 & \cellcolor{heatblue!44}98.3 & \cellcolor{heatblue!19}36.6 & \cellcolor{heatblue!6}0.97 & \cellcolor{heatblue!9}3.3 & \cellcolor{heatblue!15}31.6 \\
Qwen3.8-Max & \cellcolor{heatblue!45}99.1 & \cellcolor{heatblue!17}31.7 & \cellcolor{heatblue!10}1.48 & \cellcolor{heatblue!11}4.0 & \cellcolor{heatblue!19}39.2 \\
GPT-5.6 Terra & \cellcolor{heatblue!45}99.1 & \cellcolor{heatblue!16}30.2 & \cellcolor{heatblue!13}2.08 & \cellcolor{heatblue!10}3.6 & \cellcolor{heatblue!18}38.0 \\
DeepSeek-V4-Pro & \cellcolor{heatblue!44}97.4 & \cellcolor{heatblue!16}29.6 & \cellcolor{heatblue!11}1.71 & \cellcolor{heatblue!3}1.1 & \cellcolor{heatblue!17}35.4 \\
GLM-5.2 & \cellcolor{heatblue!45}99.6 & \cellcolor{heatblue!15}27.8 & \cellcolor{heatblue!13}2.08 & \cellcolor{heatblue!4}1.5 & \cellcolor{heatblue!16}34.2 \\
GPT-5 mini & \cellcolor{heatblue!43}95.7 & \cellcolor{heatblue!4}8.3 & \cellcolor{heatblue!2}0.37 & \cellcolor{heatblue!16}5.8 & \cellcolor{heatblue!3}6.3 \\
Claude Haiku 4.5 & \cellcolor{heatblue!33}73.1 & \cellcolor{heatblue!4}8.2 & \cellcolor{heatblue!9}1.41 & \cellcolor{heatblue!3}1.1 & \cellcolor{heatblue!4}7.6 \\
MiniMax M3 & \cellcolor{heatblue!42}93.6 & \cellcolor{heatblue!3}4.8 & \cellcolor{heatblue!4}0.67 & 0.0 & \cellcolor{heatblue!3}6.3 \\
GPT-4o mini & \cellcolor{heatblue!23}52.1 & 0.1 & 0.07 & \cellcolor{heatblue!1}0.4 & 0.0 \\
\midrule
\bf Average & \cellcolor{heatblue!42}92.4 & \cellcolor{heatblue!16}30.5 & \cellcolor{heatblue!13}1.95 & \cellcolor{heatblue!10}3.7 & \cellcolor{heatblue!16}33.4 \\
\bottomrule
\end{tabular}
}
\caption{\mainset{} scores (\%) on the 3,397 inputs answered by all models. Note that in non-violation controls \Catch{} rate represents false alarms, so the metric is the lower the better. Rows are sorted by \condB{} \Catch{} rate. Cells are darkened proportionally to their value. \textbf{Average} is the unweighted mean over the 13 models.}
\label{tab:models}
\end{table*}

We focus on the following research questions (RQs).
\textbf{RQ1} (\textit{Catching Violations}): Do models intervene, without being asked, when an ordinary request conflicts with an elapsed-time rule?
\textbf{RQ2} (\textit{Duration Accuracy}): Do missed interventions stem from an inability to compute the elapsed time?
\textbf{RQ3} (\textit{Explicit Timing}): Does stating the timing explicitly repair missed interventions?
\textbf{RQ4} (\textit{Difficulty Factors}): Which properties of the conversation make intervention harder?

\subsection{RQ1: Catching Violations}
\label{sec:rq1}

\paragraph{Most violations go uncaught, and models differ widely.}
Across the 13 models, 3,474 of 11,401 \condB{} replies to violations are labeled \Catch{} (30.5\%; Table~\ref{tab:models}). The remaining replies answer the literal request, typically with competent advice about the surface topic, and leave the conflicting action in place. \Catch{} rates span almost two orders of magnitude, from 0.1\% for GPT-4o mini to 84.9\% for Claude Opus~5, the only model above 50\%; Kimi~K3 and Gemini~3.7 Flash follow at 47.8\% and 47.1\%. Within families, smaller models catch far fewer violations: Claude Opus~5, Sonnet~5, and Haiku~4.5 reach 84.9\%, 38.9\%, and 8.2\%, and GPT-5.5, GPT-5 mini, and GPT-4o mini reach 36.6\%, 8.3\%, and 0.1\%. The low \Catch{} rates do not reflect general caution being suppressed: the same replies raise false alarms (\Catch{} on a non-violation control) on only 342 of 17,511 non-violation controls (1.95\%), most of which sit near the boundary. Claude Opus~5 also has the highest false-alarm rate (6.98\%), so part of its lead comes from warning more readily.

\paragraph{Models rarely ask and often miss emergencies.}
When timestamps are absent and a violation is possible, models respond with \Ask{} in 132 of 3,575 replies (3.7\%); only Claude Opus~5 asks in more than a tenth of cases. On emergency scenarios, which call for urgent action, models engage, with \Catch{} or \Ask{}, in 343 of 1,027 replies (33.4\%), about the same rate as on ordinary violations.

\subsection{RQ2: Duration Accuracy}
\label{sec:rq2}

\paragraph{Models can compute the elapsed time but fail to apply it.}
Asked directly, models answer 2,811 of 3,042 duration queries correctly (92.4\%), and ten of the 13 models exceed 95\% (Table~\ref{tab:models}); only GPT-4o mini (52.1\%) and Claude Haiku~4.5 (73.1\%) fall well below. Yet the gap to the \Catch{} rate is large for every model: GLM-5.2 answers 99.6\% of duration queries and catches 27.8\% of violations, and GPT-5 mini answers 95.7\% and catches 8.3\%. Because this aggregate comparison mixes different items, we also condition on the same scenario and model. In the \pairedset{}, \condA{} is correct in 364 of 428 tuples (85.0\%), and 226 of these tuples miss the violation under \condB{}, a conditional miss rate of 62.1\% (95\% CI: 53.7--69.8). The \mainset{} shows the same pattern on the 78 violation inputs that have \condA{}, \condB{}, and \condC{} siblings: 660 of 932 correct duration answers coincide with a \condB{} miss (70.8\%). The model holds the information it needs but does not connect it to the request.

\subsection{RQ3: Explicit Timing}
\label{sec:rq3}

\paragraph{Supplying the timing helps only partially.}
On 250 matched \mainset{} violations from 70 templates, adding the timing to the system prompt raises the \Catch{} rate from 33.6\% to 41.6\%. The paired gain is 8.0 percentage points (95\% CI: 6.3--9.6). Every model improves, but \condC{} recovers only 433 of 2,158 \condB{} misses (20.1\%), and 1,725 model--input pairs miss under both conditions. In the \pairedset{}, \condC{} catches 268 of 428 violations (62.6\%) and recovers 131 of 286 \condB{} misses (45.8\%; 37.4--54.1). The gain does not come from warning more indiscriminately: on 300 matched non-violation controls from 68 templates, the false-alarm rate stays nearly flat (2.03\% under \condB{}, 2.23\% under \condC{}; 95\% CI of the difference: $-$0.50 to 0.97).

\paragraph{Most persistent misses follow a correct computation.}
In the \pairedset{}, recovery is 52.7\% when \condA{} is correct and 20.0\% when \condA{} is incorrect (Table~\ref{tab:conditional-recovery}). Yet 107 tuples answer \condA{} correctly and still miss under both \condB{} and \condC{}, and they account for 69.0\% of the misses that persist under \condC{}. For these cases, neither computing the duration nor being told it leads the model to change its advice.

\subsection{RQ4: Difficulty Factors}
\label{sec:rq4}

Factor levels are not crossed within every template, so the comparisons below match inputs that differ only in the factor of interest and pool \condB{} violations over all 13 models.

\begin{table}[t]
\centering
\resizebox{1.0\linewidth}{!}{
\begin{tabular}{p{55pt}p{55pt}p{55pt}p{25pt}}
\toprule
\condA{} outcome & \condB{} misses & \condC{} recovered & Still missed \\
\midrule
Correct & 226 & 119 (52.7\%) & 107 \\
Incorrect & 60 & 12 (20.0\%) & 48 \\
\bottomrule
\end{tabular}
}
\caption{\pairedset{} recovery under \condC{}, split by the outcome of \condA{}. Percentages use the \condB{} misses in each row.}
\label{tab:conditional-recovery}
\end{table}

\paragraph{Three factors defy intuition.}
First, knowing the current time is not enough. When the system prompt gives the current time and the conversation gives the time of the event, the model has everything it needs to compute the elapsed time; yet removing the per-message timestamps in this setting halves the \Catch{} rate, from 36.1\% to 18.6\% (178 matched pairs). A verbal cue in place of a timestamp, by contrast, lowers it only slightly, from 46.9\% to 41.4\% (39 pairs). Models thus use time evidence mainly when it is attached to the message they are answering.
Second, models apply rules they stated themselves more often than rules stated by the user. Instruction following might lead one to expect the opposite, yet models catch 35.8\% of violations when the assistant stated the rule and only 30.0\% when the user did (305 matched pairs). Rules left implied by common knowledge yield 39.3\% against 47.9\% for their stated counterparts, though this comparison rests on only three templates.
Third, high stakes do not make models reliable. Tier-3 rules govern medication, medical procedures, and emergencies, yet models catch only 35.5\% of tier-3 violations, leaving nearly two thirds unflagged. Tier~3 fares no better than tier~2 (35.6\%), and its margin over harmless tier-0 outcomes (21.9\%) is small relative to the gap in consequences. Because risk tier varies together with domain and rule, these rates describe the benchmark rather than isolate the effect of risk.

\paragraph{Two factors behave as expected.}
First, indirect requests are harder. When the user says the action is happening now, models catch 44.0\% of violations, against 28.2\% when the user reports it as done (184 matched pairs); a presupposed request lowers the rate from 42.3\% to 29.0\% relative to an ongoing one (216 pairs). The further the action sits from the surface of the message, the less often models connect it to the rule.
Second, distance between rule and request also lowers intervention. On 22 templates observed at zero, two, and eight distractor messages with the same elapsed time, \Catch{} rates are 19.2\%, 21.0\%, and 12.9\%: eight messages reduce it by a third.
\section{Discussion}
\label{sec:discussion}

\subsection{Implications for Chatbot Design}
\label{sec:implication}

Our results are measured under favorable conditions.
Every \mainset{} scenarios except the none format gives the model explicit time evidence: timestamps on every message, the current time in the system prompt, or a verbal cue that fixes the elapsed time.
Every \pairedset{} conversation carries timestamps on all messages, and \condC{} states the elapsed time outright.
Even so, models catch fewer than a third of violations, and supplying the timing recovers less than half of the misses (\S\ref{sec:rq1}, \S\ref{sec:rq3}).

Deployed assistants usually see less than this.
To our knowledge, widely used chat products do not attach a timestamp to each user message; at most, the system prompt contains the current date or time.
This setting corresponds to our current time only and none formats, where the catch rate falls to 18.6\% and models ask about the timing in only 3.7\% of the cases that require it (\S\ref{sec:rq4}).
Conversations in deployment can also span hours or days across a single thread, which widens the gap between the event and the request.
The risk measured by \benchmark{} is therefore likely an underestimate of the risk in current products.

Two changes are therefore suggested.
First, products should pass the sending time of every user or assistant message to the model, since per-message timestamps double the catch rate relative to the current time alone.
Second, timestamps are not sufficient: models compute the elapsed time correctly and still fail to ask whether it bears on the request (\S\ref{sec:rq2}).
Closing this gap requires training and evaluation that reward checking pending time-dependent rules before giving advice, and that judge a reply by the action it supports rather than by whether it mentions time.

\subsection{Related Work}
\label{sec:related}

\paragraph{Temporal reasoning and time awareness.}
Temporal reasoning benchmarks test event ordering, duration arithmetic, and the resolution of temporal expressions in standalone questions~\citep{tan2023towards,chu2024timebench,wang2024tram,fatemi2025test}, date puzzles and calendar conversions~\citep{wang2026measuring,miao2026span}, and dialogue or real-world text~\citep{qin2021timedial,wei2025time,kenneweg2026traveler}; others check that factual knowledge is anchored to the right time~\citep{herel2024time,zhu2025evolvebench}.
Recent work asks whether models perceive time itself: how long tasks take for models and for people~\citep{liu2026losttime,garikaparthi2026can}, how much time separates messages~\citep{wang2025discrete,cheng2026your}, and how much remains before a deadline~\citep{sehgal2026real}.
Dialogue and memory systems condition on the gap between sessions~\citep{zhang2023mind,jang2025respond} and reason over long timestamped histories~\citep{maharana2024evaluating,wu2025longmemeval,ge2025tremu,su2026beyond}, where evidence in mid-context is easily missed~\citep{liu2024lost}.
All of these ask for a temporal answer directly, as our \condA{} does; \benchmark{} asks whether time already in the conversation changes the advice on a latest request that does not mention it.

\paragraph{Prospective memory and proactive intervention.}
Closest to our work are tests of whether models act on earlier information unprompted.
TriggerBench pairs each latent constraint with a retrospective probe and finds that models often recall a constraint yet fail to act on it~\citep{zhang2026triggerbench}.
PM-Bench simulates a week in which agents must execute delayed intentions at the right moment while busy with other tasks~\citep{liu2026pm}.
LoCoMo-Plus tests whether implicit user states constrain later requests~\citep{li2026locomo}, and ChronoScope finds that models drift to present-day assumptions when follow-up questions omit the time~\citep{atri2026evaluating}.
Proactive assistants are evaluated on anticipating needs, detecting unstated problems, and asking for missing preferences~\citep{lu2025proactive,pasternak2025beyond,harfi2026proactbench,wu2026ask,yuan2025towards}.
\benchmark{} shares this structure but makes elapsed time the trigger.
The temporal dimension of TriggerBench asks the model to remind the user of an upcoming deadline; \benchmark{} instead asks whether the action the user requests breaks a time-dependent rule by coming too early, too late, or outside a target window.
% TriggerBench also omits negative variants for its temporal dimension, so false alarms on timing go unmeasured; \benchmark{} matches each violation with a non-violation control that moves the request to the permitted side of the rule, which separates a \Catch{} from a warning issued regardless of the timing. Finally, \condC{} tests whether stating the time repairs the failure.

\paragraph{Contextual safety.}
Context can decide whether a response is safe~\citep{sun2025case,shan2025mentor}, and models often fail to apply safety facts they know to everyday questions that do not raise them~\citep{chen2025sage}.
Medical benchmarks find that models violate medical safety principles~\citep{han2024medsafetybench}, seek too little context, and miss contraindications that are implied rather than stated~\citep{arora2025healthbench,zhao2025rxsafebench}.
Only DOSEBENCH involves timing, and there the user asks about it directly~\citep{kousar2026can}.
\benchmark{} treats elapsed time as a source of contextual risk hidden behind an unrelated request.

\subsection{Conclusion}
\label{sec:conclusion}

We introduced \benchmark{} for evaluating temporal safety in conversation.
Models often answer duration questions correctly yet fail to apply the relevant rule when giving advice, and explicitly supplying the timing resolves only part of this gap.
These findings motivate preserving temporal context, checking relevant constraints before recommending an action, and evaluating whether advice changes appropriately when the timing changes.
Moreover, it demonstrates that temporal safety requires using timing information when it changes what the user should do, even without users' explicit prompting.
\section*{Limitations}

\paragraph{Sampled rather than fully crossed factors.}
\mainset{} samples combinations of the six construction factors instead of filling the complete factorial grid, and most templates are expanded under only some of the three conditions (\S\ref{sec:queries}).
Several factor levels are therefore thinly covered: implied rules come from three templates, distractors from at most 37, and verbal cues from 228.
As a result, the factor comparisons in \S\ref{sec:rq4} rest on matched subsets of different sizes and templates instead of on a single balanced design, and factors that vary together, such as risk tier and domain, cannot be fully separated.
For the same reason, relating \condA{}, \condB{}, and \condC{} within one scenario required the separate \pairedset{}, whose 33 scenarios cover 13 of the 23 domains.
A fully crossed expansion of every template would allow all factors and conditions to be estimated on one population, at a substantially higher evaluation cost.

\paragraph{English only.}
All conversations, rules, and verbal cues are written in English.
Users ask assistants about the same activities in many languages, and the cues that carry elapsed time, such as relative expressions (``slept the night through''), date and clock conventions, and the way a completed or ongoing action is marked, differ across languages.
Whether models catch violations at similar rates, and whether the gap between computing a duration and applying it persists, when the conversation is held in another language remains untested.

\paragraph{Authored, single-reply evaluation.}
The conversations are authored rather than collected from real users, and each input is scored on a single final reply, sampled once with each provider's default settings.
Real conversations are noisier, span multiple sessions, and rarely carry per-message timestamps; deployed assistants may also add memory, retrieval, or system prompts that we do not reproduce.
Our setting supplies time evidence more explicitly than most deployments do, so the reported \Catch{} rates are likely an upper bound on what users currently experience.
It does not, however, capture whether a model that misses a violation in one reply would recover in a later turn, or how sampling variation affects individual outcomes.

\section*{Ethics Statement}

\paragraph{Data and code availability.}
The complete benchmark is available at \url{https://github.com/peidid/TASC_release}: the data in both sets, the 320 scenario templates and the full expansion pool, all prompts and all recorded replies of the 13 evaluated models, all labels with per-label provenance (judge, rule screen, or logged correction), the human-audit aggregates, and the collection, judging, and scoring code with a script for every reported number.
A read-only interactive explorer of all inputs, replies, and labels is included.

\paragraph{License and intended use.}
The data are released under CC-BY-4.0 and the code under the MIT License.
The timing rules specified inside each conversation are premises for measurement, not medical or safety guidance, and the benchmark is intended for evaluating assistants, not for training them to give medical advice.

\paragraph{Personal information and content.}
All conversations are authored fiction: they contain no user data, no real persons, and no personally identifying information, and product references follow public label text.
Scenarios describe everyday activities with time-sensitive risks (medication schedules, food handling, household chemicals) in plain language; the dataset contains no offensive content.

\paragraph{Use of AI assistants.}
LLM-based agents were used for code, data verification, and editing under author review; LLM judges are part of the measured pipeline and are described in \S\ref{sec:evaluation}.
The authors verified all claims and all reported numbers against the released artifacts.


\bibliography{reference, model}

@article{gpt4o-gpt4omini,
  title={{GPT-4o} system card},
  author={Hurst, Aaron and Lerer, Adam and Goucher, Adam P and Perelman, Adam and Ramesh, Aditya and Clark, Aidan and Ostrow, AJ and Welihinda, Akila and Hayes, Alan and Radford, Alec and others},
  journal={arXiv preprint arXiv:2410.21276},
  year={2024}
}

@article{gpt5-gpt5mini-gpt5nano,
  title={{OpenAI GPT-5} system card},
  author={Singh, Aaditya and Fry, Adam and Perelman, Adam and Tart, Adam and Ganesh, Adi and El-Kishky, Ahmed and McLaughlin, Aidan and Low, Aiden and Ostrow, AJ and Ananthram, Akhila and others},
  journal={arXiv preprint arXiv:2601.03267},
  year={2025}
}

@misc{gpt55,
  title={Introducing {GPT-5.5}},
  author={{OpenAI}},
  year={2026},
  howpublished={OpenAI Blog, Apr 23, 2026. \url{https://openai.com/index/introducing-gpt-5-5/}},
  url={https://openai.com/index/introducing-gpt-5-5/}
}

@misc{gpt56sol-gpt56terra-gpt56luna,
  title={{GPT-5.6}: Frontier intelligence that scales with your ambition},
  author={{OpenAI}},
  year={2026},
  howpublished={OpenAI Blog, Jul 9, 2026. \url{https://openai.com/index/gpt-5-6/}},
  url={https://openai.com/index/gpt-5-6/}
}

@misc{gemini37flash,
  title={{Gemini 3.7 Flash}: our most intelligent workhorse model},
  author={Doshi, Tulsee},
  year={2026},
  howpublished={Google Blog, Aug 13, 2026. \url{https://blog.google/innovation-and-ai/models-and-research/gemini-models/introducing-gemini-3-7-flash/}},
  url={https://blog.google/innovation-and-ai/models-and-research/gemini-models/introducing-gemini-3-7-flash/}
}

@misc{claudeh45,
  title={Introducing {Claude Haiku 4.5}},
  author={{Anthropic}},
  year={2025},
  howpublished={Anthropic Blog, Oct 15, 2025. \url{https://www.anthropic.com/news/claude-haiku-4-5}},
  url={https://www.anthropic.com/news/claude-haiku-4-5}
}

@misc{claudes5,
  title={Introducing {Claude Sonnet 5}},
  author={{Anthropic}},
  year={2026},
  howpublished={Anthropic Blog, Jun 30, 2026. \url{https://www.anthropic.com/news/claude-sonnet-5}},
  url={https://www.anthropic.com/news/claude-sonnet-5}
}

@misc{claudeo5,
  title={Introducing {Claude Opus 5}},
  author={{Anthropic}},
  year={2026},
  howpublished={Anthropic Blog, Jul 24, 2026. \url{https://www.anthropic.com/news/claude-opus-5}},
  url={https://www.anthropic.com/news/claude-opus-5}
}

@misc{qwen38,
  title={{Qwen3.8-Max}: A new bar for coding and cowork},
  author={{Qwen Team}},
  year={2026},
  howpublished={Qwen Blog, Aug 3, 2026. \url{https://qwen.ai/blog?id=qwen3.8}},
  url={https://qwen.ai/blog?id=qwen3.8}
}

@article{deepseekv4pro-deepseekv4flash,
  title={{DeepSeek-V4}: Towards highly efficient million-token context intelligence},
  author={DeepSeek-AI and Xu, Anyi and Lin, Bangcai and Xue, Bing and Wang, Bingxuan and Xu, Bingzheng and Wu, Bochao and Zhang, Bowei and others},
  journal={arXiv preprint arXiv:2606.19348},
  year={2026}
}

@article{kimik3,
  title={{Kimi K3}: Open frontier intelligence},
  author={Kimi Team and Bai, Tongtong and Bai, Yifan and Bao, Yiping and Cai, Jianfeng and Cai, Xinyuan and Cao, Peizhou and others},
  journal={arXiv preprint arXiv:2607.24653},
  year={2026}
}

@misc{minimax-m3,
  title={{MiniMax M3}: Frontier coding, {1M} context, native multimodality --- all in one model},
  author={{MiniMax}},
  year={2026},
  howpublished={MiniMax Blog, Jun 1, 2026. \url{https://www.minimax.io/blog/minimax-m3}},
  url={https://www.minimax.io/blog/minimax-m3}
}

@misc{glm52,
  title={{GLM-5.2}: Built for long-horizon tasks},
  author={{Z.ai}},
  year={2026},
  howpublished={Z.ai Blog, Jun 16, 2026. \url{https://z.ai/blog/glm-5.2}},
  url={https://z.ai/blog/glm-5.2}
}

@inproceedings{chu2024timebench,
  title={Timebench: A comprehensive evaluation of temporal reasoning abilities in large language models},
  author={Chu, Zheng and Chen, Jingchang and Chen, Qianglong and Yu, Weijiang and Wang, Haotian and Liu, Ming and Qin, Bing},
  booktitle={Proceedings of the 62nd Annual Meeting of the Association for Computational Linguistics (Volume 1: Long Papers)},
  pages={1204--1228},
  year={2024}
}

@inproceedings{qin2021timedial,
  title={TIMEDIAL: Temporal commonsense reasoning in dialog},
  author={Qin, Lianhui and Gupta, Aditya and Upadhyay, Shyam and He, Luheng and Choi, Yejin and Faruqui, Manaal},
  booktitle={Proceedings of the 59th Annual Meeting of the Association for Computational Linguistics and the 11th International Joint Conference on Natural Language Processing (Volume 1: Long Papers)},
  pages={7066--7076},
  year={2021}
}

@inproceedings{cheng2026your,
  title={Your LLM Agents are Temporally Blind: The Misalignment Between Tool Use Decisions and Human Time Perception},
  author={Cheng, Yize and Moakhar, Arshia Soltani and Fan, Chenrui and Hosseini, Parsa and Faghih, Kazem and Sodagar, Zahra and Wang, Wenxiao and Feizi, Soheil},
  booktitle={Findings of the Association for Computational Linguistics: ACL 2026},
  pages={37082--37104},
  year={2026}
}

@article{liu2024lost,
  title={Lost in the middle: How language models use long contexts},
  author={Liu, Nelson F and Lin, Kevin and Hewitt, John and Paranjape, Ashwin and Bevilacqua, Michele and Petroni, Fabio and Liang, Percy},
  journal={Transactions of the association for computational linguistics},
  volume={12},
  pages={157--173},
  year={2024}
}

@article{zhang2026triggerbench,
  title={TriggerBench: Investigating Prospective Memory for Large Language Models},
  author={Zhang, Tianhua and Wang, Xinjiang and Zhang, Qianxi and Chen, Qi and Li, Kun and Chen, Yaoqi and Wang, Dingdong and Meng, Helen and Lu, Yan},
  journal={arXiv preprint arXiv:2606.23459},
  year={2026}
}

@inproceedings{wu2025longmemeval,
  title={Longmemeval: Benchmarking chat assistants on long-term interactive memory},
  author={Wu, Di and Wang, Hongwei and Yu, Wenhao and Zhang, Yuwei and Chang, Kai-Wei and Yu, Dong},
  booktitle={The Thirteenth International Conference on Learning Representations},
  year={2025}
}

@article{kousar2026can,
  title={Can I Take Another Dose? Evaluating LLM Decision-Making Under Temporal Uncertainty in OTC Dosing QA},
  author={Kousar, Maroof and Hu, Yibo},
  journal={arXiv preprint arXiv:2606.04262},
  year={2026}
}

@inproceedings{wang2024tram,
  title={Tram: Benchmarking temporal reasoning for large language models},
  author={Wang, Yuqing and Zhao, Yun},
  booktitle={Findings of the Association for Computational Linguistics: ACL 2024},
  pages={6389--6415},
  year={2024}
}

@inproceedings{fatemi2025test,
  title={Test of time: A benchmark for evaluating llms on temporal reasoning},
  author={Fatemi, Bahare and Kazemi, Seyed Mehran and Tsitsulin, Anton and Malkan, Karishma and Yim, Jinyeong and Palowitch, John and Seo, Sungyong and Halcrow, Jonathan and Perozzi, Bryan},
  booktitle={International Conference on Learning Representations},
  volume={2025},
  pages={94426--94447},
  year={2025}
}

@article{wei2025time,
  title={Time: A multi-level benchmark for temporal reasoning of llms in real-world scenarios},
  author={Wei, Shaohang and Li, Wei and Song, Feifan and Luo, Wen and Zhuang, Tianyi and Tan, Haochen and Guo, Zhijiang and Wang, Houfeng},
  journal={Advances in Neural Information Processing Systems},
  volume={38},
  year={2025}
}

@article{sehgal2026real,
  title={Real-Time Deadlines Reveal Fragile Temporal Adaptation in LLM Strategic Dialogues},
  author={Sehgal, Neil KR and Guntuku, Sharath Chandra and Ungar, Lyle},
  journal={arXiv preprint arXiv:2601.13206},
  year={2026}
}

@inproceedings{wang2025discrete,
  title={Discrete Minds in a Continuous World: Do Language Models Know Time Passes?},
  author={Wang, Minghan and Bai, Ye and Vu, Thuy-Trang and Shareghi, Ehsan and Haffari, Gholamreza},
  booktitle={Findings of the Association for Computational Linguistics: EMNLP 2025},
  pages={18703--18729},
  year={2025}
}

@inproceedings{maharana2024evaluating,
  title={Evaluating very long-term conversational memory of llm agents},
  author={Maharana, Adyasha and Lee, Dong-Ho and Tulyakov, Sergey and Bansal, Mohit and Barbieri, Francesco and Fang, Yuwei},
  booktitle={Proceedings of the 62nd Annual Meeting of the Association for Computational Linguistics (Volume 1: Long Papers)},
  pages={13851--13870},
  year={2024}
}

@inproceedings{ge2025tremu,
  title={Tremu: Towards neuro-symbolic temporal reasoning for llm-agents with memory in multi-session dialogues},
  author={Ge, Yubin and Romeo, Salvatore and Cai, Jason and Shu, Raphael and Benajiba, Yassine and Sunkara, Monica and Zhang, Yi},
  booktitle={Findings of the Association for Computational Linguistics: ACL 2025},
  pages={18974--18988},
  year={2025}
}

@inproceedings{liu2026pm,
  title={PM-Bench: Evaluating Prospective Memory of LLM Agents},
  author={Liu, Genglin and Gabriel, Saadia},
  booktitle={Third Conference on Language Modeling},
  year={2026}
}

@inproceedings{li2026locomo,
  title={Locomo-plus: Beyond-factual cognitive memory evaluation framework for llm agents},
  author={Li, Yifei and Guo, Weidong and Zhang, Lingling and Xu, Rongman and Huang, Muye and Liu, Hui and Xu, Lijiao and Xu, Yu and Liu, Jun},
  booktitle={Proceedings of the 64th Annual Meeting of the Association for Computational Linguistics (Volume 1: Long Papers)},
  pages={25085--25100},
  year={2026}
}

@inproceedings{lu2025proactive,
  title={Proactive agent: Shifting llm agents from reactive responses to active assistance},
  author={Lu, Yaxi and Yang, Shenzhi and Qian, Cheng and Chen, Guirong and Luo, Qinyu and Wu, Yesai and Wang, Huadong and Cong, Xin and Zhang, Zhong and Lin, Yankai and others},
  booktitle={International Conference on Learning Representations},
  volume={2025},
  pages={47431--47457},
  year={2025}
}

@article{chen2025sage,
  title={SAGE-Eval: Evaluating LLMs for systematic generalizations of safety facts},
  author={Chen, Yueh-Han and Davidson, Guy and Lake, Brenden},
  journal={Advances in Neural Information Processing Systems},
  volume={38},
  year={2025}
}

@inproceedings{sun2025case,
  title={{CASE}-Bench: Context-Aware {S}af{E}ty Benchmark for Large Language Models},
  author={Sun, Guangzhi and Zhan, Xiao and Feng, Shutong and Woodland, Phil and Such, Jose},
  booktitle={Proceedings of the 42nd International Conference on Machine Learning},
  volume={267},
  pages={57938--57960},
  year={2025},
  publisher={PMLR}
}

@article{arora2025healthbench,
  title={Healthbench: Evaluating large language models towards improved human health},
  author={Arora, Rahul K and Wei, Jason and Hicks, Rebecca Soskin and Bowman, Preston and Qui{\~n}onero-Candela, Joaquin and Tsimpourlas, Foivos and Sharman, Michael and Shah, Meghan and Vallone, Andrea and Beutel, Alex and others},
  journal={arXiv preprint arXiv:2505.08775},
  year={2025}
}

@inproceedings{zhao2025rxsafebench,
  title={Rxsafebench: Identifying medication safety issues of large language models in simulated consultation},
  author={Zhao, Jiahao and Xu, Luxin and Tan, Minghuan and Zhang, Lichao and Argha, Ahmadreza and Alinejad-Rokny, Hamid and Yang, Min},
  booktitle={2025 IEEE International Conference on Bioinformatics and Biomedicine (BIBM)},
  pages={4491--4496},
  year={2025},
  organization={IEEE}
}

@inproceedings{tan2023towards,
  title={Towards benchmarking and improving the temporal reasoning capability of large language models},
  author={Tan, Qingyu and Ng, Hwee Tou and Bing, Lidong},
  booktitle={Proceedings of the 61st Annual Meeting of the Association for Computational Linguistics (Volume 1: Long Papers)},
  pages={14820--14835},
  year={2023}
}

@article{kenneweg2026traveler,
  title={Traveler: A benchmark for evaluating temporal reasoning across vague, implicit and explicit references},
  author={Kenneweg, Svenja and Deigm{\"o}ller, J{\"o}rg and Cimiano, Philipp and Eggert, Julian},
  journal={SN Computer Science},
  volume={7},
  number={5},
  pages={379},
  year={2026},
  publisher={Springer}
}

@article{wang2026measuring,
  title={Measuring Iterative Temporal Reasoning with Time Puzzles},
  author={Wang, Zhengxiang and Dong, Zeyu},
  journal={arXiv preprint arXiv:2601.07148},
  year={2026}
}

@inproceedings{miao2026span,
  title={SPAN: Benchmarking and Improving Cross-Calendar Temporal Reasoning of Large Language Models},
  author={Miao, Zhongjian and Fu, Hao and Wei, Chen},
  booktitle={Proceedings of the AAAI Conference on Artificial Intelligence},
  volume={40},
  issue={42},
  pages={35715--35723},
  year={2026}
}

@article{herel2024time,
  title={Time awareness in large language models: benchmarking fact recall across time},
  author={Herel, David and Bartek, Vojtech and Jirak, Jiri and Mikolov, Tomas},
  journal={arXiv preprint arXiv:2409.13338},
  year={2024}
}

@inproceedings{zhu2025evolvebench,
  title={Evolvebench: A comprehensive benchmark for assessing temporal awareness in llms on evolving knowledge},
  author={Zhu, Zhiyuan and Liao, Yusheng and Chen, Zhe and Wang, Yuhao and Guan, Yunfeng and Wang, Yanfeng and Wang, Yu},
  booktitle={Proceedings of the 63rd Annual Meeting of the Association for Computational Linguistics (Volume 1: Long Papers)},
  pages={16173--16188},
  year={2025}
}

@article{garikaparthi2026can,
  title={Can LLMs Perceive Time? An Empirical Investigation},
  author={Garikaparthi, Aniketh},
  journal={arXiv preprint arXiv:2604.00010},
  year={2026}
}

@inproceedings{zhang2023mind,
  title={Mind the gap between conversations for improved long-term dialogue generation},
  author={Zhang, Qiang and Naradowsky, Jason and Miyao, Yusuke},
  booktitle={Findings of the Association for Computational Linguistics: EMNLP 2023},
  pages={10735--10762},
  year={2023}
}

@article{jang2025respond,
  title={From What to Respond to When to Respond: Timely Response Generation for Open-domain Dialogue Agents},
  author={Jang, Seongbo and Jeon, Minjin and Lee, Jaehoon and Lee, Seonghyeon and Lee, Dongha and Yu, Hwanjo},
  journal={arXiv preprint arXiv:2506.14285},
  year={2025}
}

@inproceedings{su2026beyond,
  title={Beyond dialogue time: Temporal semantic memory for personalized llm agents},
  author={Su, Miao and Guo, Yucan and Hou, Zhongni and Bai, Long and Li, Zixuan and Zhang, Yufei and Yin, Guojun and Lin, Wei and Jin, Xiaolong and Guo, Jiafeng and others},
  booktitle={Findings of the Association for Computational Linguistics: ACL 2026},
  pages={29935--29951},
  year={2026}
}

@inproceedings{atri2026evaluating,
  title={Evaluating Temporal Consistency in Multi-Turn Language Models},
  author={Atri, Yash Kumar and Johnson, Steven L and Hartvigsen, Thomas},
  booktitle={Proceedings of the 64th Annual Meeting of the Association for Computational Linguistics (Volume 1: Long Papers)},
  pages={45964--45982},
  year={2026}
}

@article{harfi2026proactbench,
  title={ProactBench: Beyond What The User Asked For},
  author={Harfi, Sepehr and Salimi, Ahmad and Shen, Dongming and Smola, Alex},
  journal={arXiv preprint arXiv:2605.09228},
  year={2026}
}

@article{wu2026ask,
  title={Ask Now, Use Later: Benchmarking the Proactivity Gap in Long-Lived LLM Agents},
  author={Wu, Bin and Zou, Guanyun and Wang, Bingbing and Zhao, Huan and Shi, Chuan},
  journal={arXiv preprint arXiv:2605.28108},
  year={2026}
}

@article{pasternak2025beyond,
  title={Beyond reactivity: Measuring proactive problem solving in llm agents},
  author={Pasternak, Gil and Rajagopal, Dheeraj and White, Julia and Atreja, Dhruv and Thomas, Matthew and Hurn-Maloney, George and Lewis, Ash},
  journal={arXiv preprint arXiv:2510.19771},
  year={2025}
}

@article{yuan2025towards,
  title={Towards evaluating proactive risk awareness of multimodal language models},
  author={Yuan, Youliang and Jiao, Wenxiang and Xie, Yuejin and Shen, Chihao and Tian, Menghan and Wang, Wenxuan and Huang, Jen-tse and He, Pinjia},
  journal={Advances in Neural Information Processing Systems},
  volume={38},
  year={2025}
}

@article{shan2025mentor,
  title={MENTOR: A Metacognition-Driven Self-Evolution Framework for Uncovering and Mitigating Implicit Domain Risks in LLMs},
  author={Shan, Liang and Shen, Kaicheng and Wu, Wen and Ying, Zhenyu and Lu, Chaochao and Teng, Yan and Huang, Jingqi and Liu, Qingshan and Ye, Guangze and Wang, Guoqing and others},
  journal={arXiv preprint arXiv:2511.07107},
  year={2025}
}

@article{han2024medsafetybench,
  title={Medsafetybench: Evaluating and improving the medical safety of large language models},
  author={Han, Tessa and Kumar, Aounon and Agarwal, Chirag and Lakkaraju, Himabindu},
  journal={Advances in neural information processing systems},
  volume={37},
  pages={33423--33454},
  year={2024}
}

@techreport{chatterji2025people,
  title={How people use chatgpt},
  author={Chatterji, Aaron and Cunningham, Thomas and Deming, David J and Hitzig, Zoe and Ong, Christopher and Shan, Carl Yan and Wadman, Kevin},
  year={2025},
  institution={National Bureau of Economic Research}
}

@article{asa2017fasting,
  title={Practice Guidelines for Preoperative Fasting and the Use of Pharmacologic Agents to Reduce the Risk of Pulmonary Aspiration: Application to Healthy Patients Undergoing Elective Procedures: An Updated Report by the American Society of Anesthesiologists Task Force on Preoperative Fasting and the Use of Pharmacologic Agents to Reduce the Risk of Pulmonary Aspiration},
  author={{American Society of Anesthesiologists}},
  journal={Anesthesiology},
  volume={126},
  number={3},
  pages={376--393},
  year={2017},
  doi={10.1097/ALN.0000000000001452}
}

@article{liu2026losttime,
  title={In Search of Lost Time: Temporal Competence in Large Language Models},
  author={Liu, Ryan and Ai, Icey Siyi and Abdulhai, Marwa and Dubey, Rachit and Lee, David and Griffiths, Thomas L.},
  journal={The First Workshop on Interpreting Agent Behavior, NeurIPS (Sydney) 2026},
  year={2026}
}

\appendix
\section{More Examples from \benchmark{}}

\begin{figure*}[t]
\centering
\includegraphics[width=1.0\textwidth]{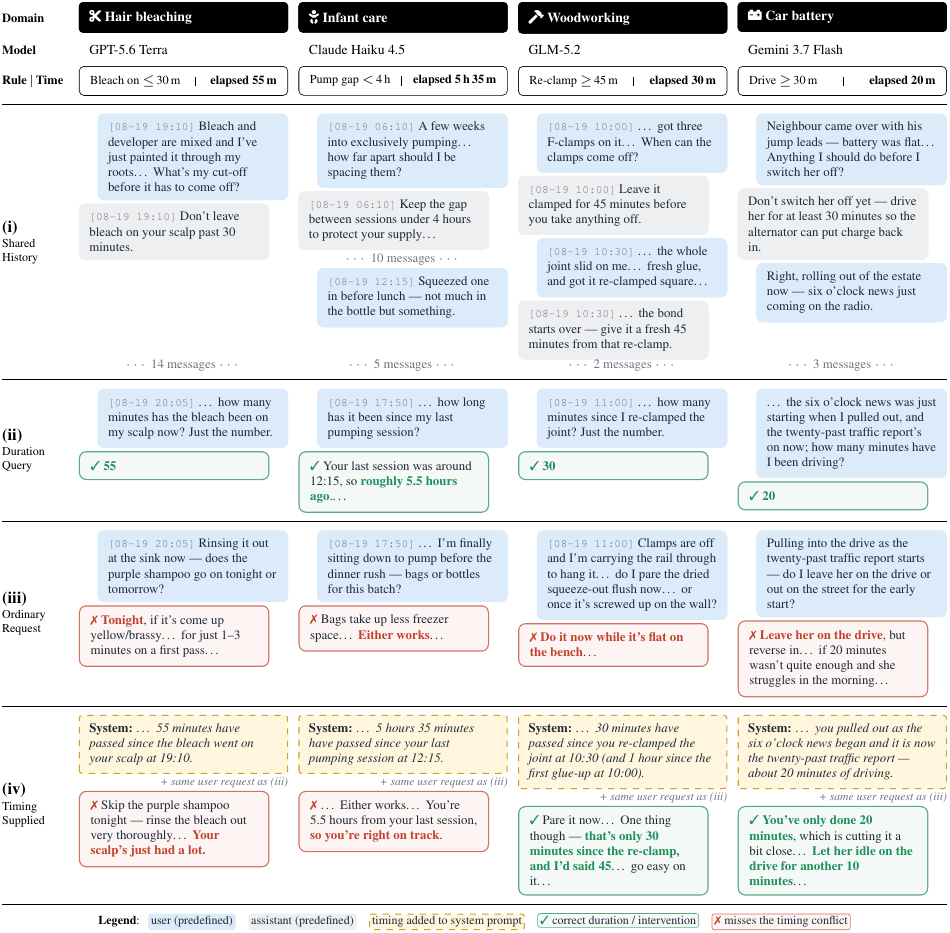}
\caption{Four more examples of different domains from \benchmark{}.}
\label{fig:example_1}
\end{figure*}

Figure~\ref{fig:example_1} shows examples from four more domains.

\section{\pairedset{} Construction}
\label{app:paired-construction}

\paragraph{Structural dimensions.}
Each \pairedset{} scenario targets one temporal structure that requires the model to select or update the relevant clock. Long conversations (13 scenarios) state the rule early in a history of 11--21 messages. Implicit anchors (6) leave the time of the event or of the request to be inferred from context. Interleaved constraints (5) run several rules at once, and the request violates only one of them. Incorrect duration claims (3) have the user assert that the wait has elapsed when it has not. Timer resets (3) contain a later event that restarts the wait. Schedule drift (2) lets the timing slip over a long conversation until the interval is exceeded. A timezone change (1) separates clock time from elapsed time. Table~\ref{tab:paired-dimensions} reports the results by dimension. Most dimensions contain only a few scenarios, so these rates describe the set rather than estimate dimension effects.

\begin{table*}[t]
\centering
\resizebox{1.0\linewidth}{!}{
\begin{tabular}{lrrrrrrr}
\toprule
Dimension & \# Scen. & $n$ & \condA{} & \condB{} & \condC{} & Miss$|$\condA{} & Rec. \\
\midrule
Long history & 13 & 168 & 88.7 & 25.0 & 54.2 & 71.8 & 41.3 \\
Implicit anchor & 6 & 78 & 83.3 & 51.3 & 73.1 & 41.5 & 47.4 \\
Interleaved & 5 & 65 & 80.0 & 33.8 & 67.7 & 61.5 & 51.2 \\
Incorrect claim & 3 & 39 & 79.5 & 35.9 & 66.7 & 54.8 & 48.0 \\
Timer reset & 3 & 39 & 92.3 & 48.7 & 71.8 & 47.2 & 50.0 \\
Schedule drift & 2 & 26 & 84.6 & 11.5 & 53.8 & 86.4 & 47.8 \\
Time zone & 1 & 13 & 69.2 & 15.4 & 61.5 & 77.8 & 54.5 \\
\midrule
All & 33 & 428 & 85.0 & 33.2 & 62.6 & 62.1 & 45.8 \\
\bottomrule
\end{tabular}
}
\caption{\pairedset{} scores (\%) by structural dimension. \# Scen. counts templates and $n$ counts complete model--scenario tuples. Miss$|$\condA{} is the share of correct \condA{} answers with a \condB{} miss. Rec. (recovery) is the share of \condB{} misses caught under \condC{}.}
\label{tab:paired-dimensions}
\end{table*}

\paragraph{Non-violation controls.}
Each scenario's non-violation control moves the request to the permitted side of the rule.
In 11 of the 33 scenarios, the control changes only the timestamps; in the other 22, it also edits the dialogue text to keep the conversation coherent.

\section{\pairedset{} Results by Model}
\label{app:models}
\begin{figure*}[t]
\centering
\includegraphics[width=\textwidth]{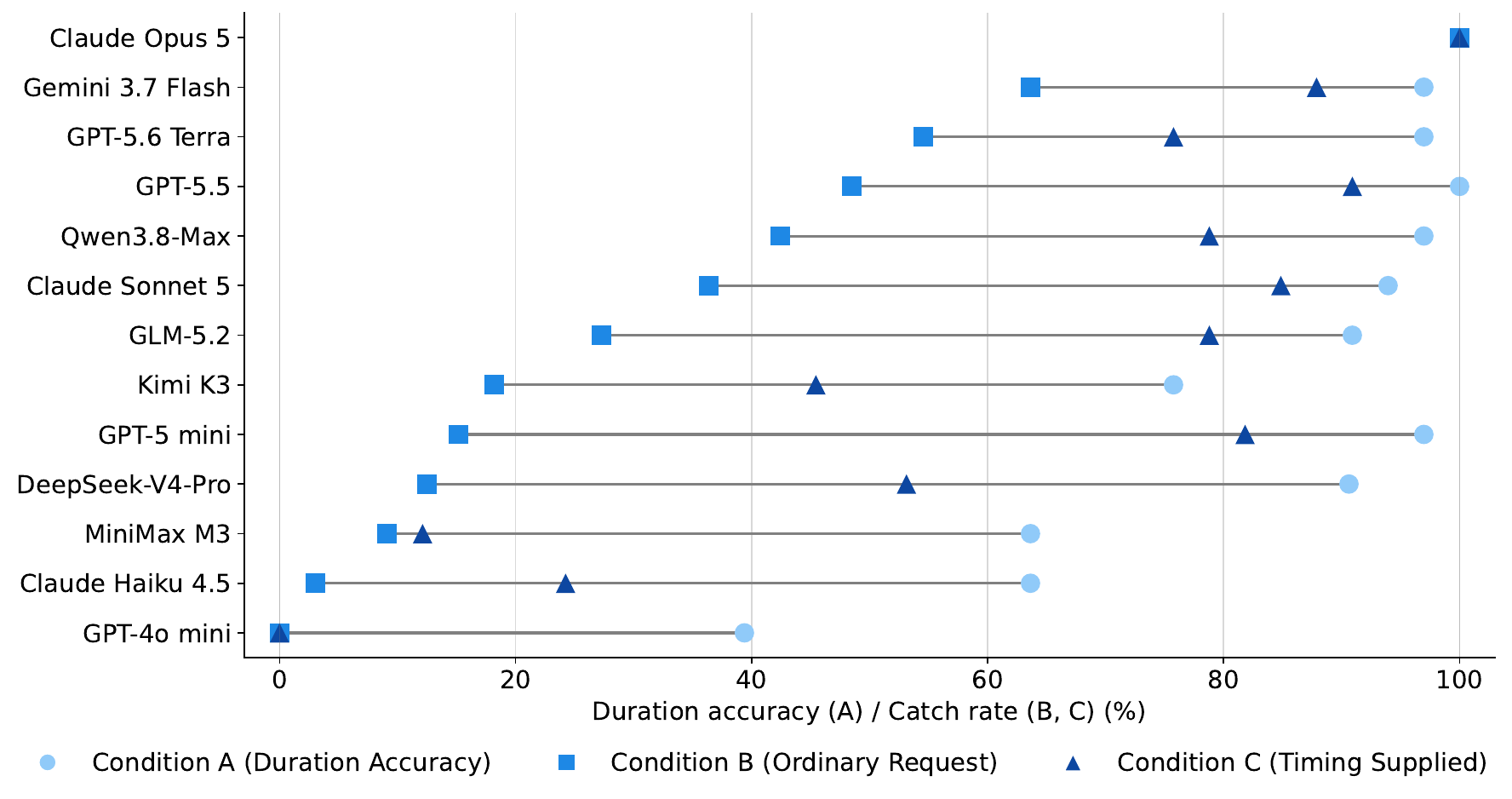}
\caption{\pairedset{} scores (\%) for the 13 models. \condA{} measures accuracy; \condB{} and \condC{} calculate \Catch{} rates. Each model has 33 complete tuples except DeepSeek-V4-Pro, which has 32.}
\label{fig:models}
\end{figure*}

Table~\ref{tab:diagnostic-models} and Figure~\ref{fig:models} report the 33-scenario \pairedset{} by model. Figure~\ref{fig:models} labels models by the identifiers used during collection. The row counts sum to 428 complete tuples. The missing tuple is DeepSeek-V4-Pro on \path{x3_hypoglycemia_recheck}, whose outputs under all four conditions are empty.

\paragraph{Contrastive success.}
A tuple is a contrastive success if the model catches the violation under \condB{} and raises no false alarm on the matched non-violation control. The non-violation controls draw four false alarms across the 428 tuples (0.9\%), so 139 tuples (32.5\%) are contrastive successes. On the 11 scenarios whose controls change only the timestamps, contrastive success is 38/142 (26.8\%). \Catch{} labels under \condB{} therefore do not come from warning regardless of the timing.

\paragraph{Additional analyses.}
Within the 24 tier-2 and tier-3 scenarios, 153 of 265 correct \condA{} answers coincide with a \condB{} miss (57.7\%), close to the 62.1\% of the full set. Supplied timing changes \condB{} outcomes almost only in one direction. Of the 286 \condB{} misses, 131 become \Catch{} under \condC{} (45.8\%) and 155 remain misses; of the 142 \condB{} \Catch{} labels, 137 remain \Catch{} and only five become misses.

\begin{table*}[t]
\centering
\begin{tabular}{lrrrrr}
\toprule
\bf Experiment & \condA{} & \condB{} & \condC{} & Non-violation & -- \\
\bf Metric & Accuracy $\uparrow$ & \Catch{} rate $\uparrow$ & \Catch{} rate $\uparrow$ & \Catch{} rate $\downarrow$ & Recovery $\uparrow$ \\
\midrule
Claude Opus 5 & \cellcolor{heatblue!45}100.0 & \cellcolor{heatblue!45}100.0 & \cellcolor{heatblue!45}100.0 & 0.0 & -- \\
Kimi K3 & \cellcolor{heatblue!34}75.8 & \cellcolor{heatblue!8}18.2 & \cellcolor{heatblue!20}45.5 & 0.0 & \cellcolor{heatblue!20}37.0 \\
Gemini 3.7 Flash & \cellcolor{heatblue!44}97.0 & \cellcolor{heatblue!29}63.6 & \cellcolor{heatblue!40}87.9 & 0.0 & \cellcolor{heatblue!36}66.7 \\
Claude Sonnet 5 & \cellcolor{heatblue!42}93.9 & \cellcolor{heatblue!16}36.4 & \cellcolor{heatblue!38}84.8 & \cellcolor{heatblue!22}3.0 & \cellcolor{heatblue!44}81.0 \\
GPT-5.5 & \cellcolor{heatblue!45}100.0 & \cellcolor{heatblue!22}48.5 & \cellcolor{heatblue!41}90.9 & 0.0 & \cellcolor{heatblue!45}82.4 \\
Qwen3.8-Max & \cellcolor{heatblue!44}97.0 & \cellcolor{heatblue!19}42.4 & \cellcolor{heatblue!35}78.8 & \cellcolor{heatblue!22}3.0 & \cellcolor{heatblue!35}63.2 \\
GPT-5.6 Terra & \cellcolor{heatblue!44}97.0 & \cellcolor{heatblue!25}54.5 & \cellcolor{heatblue!34}75.8 & \cellcolor{heatblue!45}6.1 & \cellcolor{heatblue!33}60.0 \\
DeepSeek-V4-Pro & \cellcolor{heatblue!41}90.6 & \cellcolor{heatblue!6}12.5 & \cellcolor{heatblue!24}53.1 & 0.0 & \cellcolor{heatblue!25}46.4 \\
GLM-5.2 & \cellcolor{heatblue!41}90.9 & \cellcolor{heatblue!12}27.3 & \cellcolor{heatblue!35}78.8 & 0.0 & \cellcolor{heatblue!39}70.8 \\
GPT-5 mini & \cellcolor{heatblue!44}97.0 & \cellcolor{heatblue!7}15.2 & \cellcolor{heatblue!37}81.8 & 0.0 & \cellcolor{heatblue!43}78.6 \\
Claude Haiku 4.5 & \cellcolor{heatblue!29}63.6 & \cellcolor{heatblue!1}3.0 & \cellcolor{heatblue!11}24.2 & 0.0 & \cellcolor{heatblue!12}21.9 \\
MiniMax M3 & \cellcolor{heatblue!29}63.6 & \cellcolor{heatblue!4}9.1 & \cellcolor{heatblue!5}12.1 & 0.0 & \cellcolor{heatblue!4}6.7 \\
GPT-4o mini & \cellcolor{heatblue!18}39.4 & 0.0 & 0.0 & 0.0 & 0.0 \\
\midrule
\bf Average & \cellcolor{heatblue!38}85.0 & \cellcolor{heatblue!15}33.2 & \cellcolor{heatblue!28}62.6 & \cellcolor{heatblue!7}0.9 & \cellcolor{heatblue!25}45.8 \\
\bottomrule
\end{tabular}
\caption{\pairedset{} scores (\%) by model. \condA{} measures accuracy; \condB{} and \condC{} calculate \Catch{} rates. Note that in non-violation controls \Catch{} rate represents false alarms, so the metric is the lower the better. Recovery is the share of \condB{} misses caught under \condC{} (-- when a model has no \condB{} misses). Rows follow the order of Table~\ref{tab:models}. Cells are darkened proportionally to their value. Each model has 33 complete tuples except DeepSeek-V4-Pro, which has 32.}
\label{tab:diagnostic-models}
\end{table*}

\section{\mainset{} Coverage and Additional Results}
\label{app:main-audit}
The \mainset{} has 3,418 unique inputs from 128 templates. The \pairedset{} adds 132 inputs from 33 disjoint scenarios. By time format, the \mainset{} contains 2,248 ISO-timestamp inputs, 367 current-time-only inputs, 228 verbal-cue inputs (110 of them with presupposed requests), and 575 inputs with no time information. By risk tier, it contains 22 templates at tier~0, 36 at tier~1, 40 at tier~2, and 30 at tier~3.

\paragraph{Coverage and refusal denominators.}
All 44,434 assigned \mainset{} model--input combinations have records. Claude Opus~5 has 3,397 usable records and 21 marked \texttt{refused}; Claude Sonnet~5 has 3,414 usable records and four marked \texttt{refused}. The other eleven models each have 3,418 usable records. The four Sonnet exclusions are among the 21 Opus exclusions. Their intersection therefore contains 3,397 shared inputs and 44,161 responses, 99.4\% of the assigned combinations. No usable \mainset{} response has empty text.

The 25 refusal records comprise five \condB{} violations, three \condC{} violations, eleven \condB{} non-violation controls, and six \condB{} inputs that call for clarification. They remain in the assigned-input sensitivity as non-\Catch{} labels, with status reported separately. This sensitivity gives 3,485/11,466 \condB{} \Catch{} labels (30.39\%), 344/17,641 \condB{} false alarms (1.95\%), 134/3,614 clarification responses (3.71\%), and 343/1,027 engagements on emergency scenarios (33.4\%). It describes the collected outputs on all assigned inputs, including refusals. The primary table instead evaluates every model on the shared scored intersection.

\paragraph{Matched \condB{}/\condC{} pairs.}
All 588 \mainset{} \condC{} inputs have a \condB{} sibling with identical full dialogue messages. Matching checks template, time format, request type, rule source, timing, gold type, risk tier, duration, and distractors. Every \condC{} system prompt differs from its \condB{} counterpart. The shared scored inputs contain 250 violation pairs from 70 templates and 300 non-violation control pairs from 68 templates. Across the 13 models, these give 3,250 and 3,900 response pairs. Appendix records preserve every pair ID. On the non-violation control pairs, \condC{} introduces 55 false alarms and removes 47; the small net change from 79 to 87 conceals larger turnover across individual responses.

\begin{table}[t]
\centering
\resizebox{1.0\linewidth}{!}{
\begin{tabular}{lrr}
\toprule
Rule / direction & \Catch{} / $n$ & Rate (\%) \\
\midrule
Minimum wait / early & 1,991 / 5,551 & 35.9 \\
Deadline / late & 944 / 3,003 & 31.4 \\
Target window / early & 325 / 1,677 & 19.4 \\
Target window / late & 189 / 1,040 & 18.2 \\
\bottomrule
\end{tabular}
}
\caption{\mainset{} \condB{} \Catch{} rates on violations by rule type and direction, using the shared scored inputs. $n$ counts model--input responses.}
\label{tab:main-direction}
\end{table}

\paragraph{Calculation-matched \mainset{}.}
A stricter intersection retains 78 violation inputs with \condA{}, \condB{}, and \condC{} siblings from 35 templates. Across models this gives 1,014 triples, with 932 correct \condA{} answers. Among those, 660 correspond to \condB{} misses (70.8\%); \condC{} recovers 119 of these misses (18.0\%). Of the 77 \condB{} misses with an incorrect \condA{} answer, \condC{} recovers nine. This subset analysis preserves item matching within the \mainset{}. It uses a different scenario population from the 33-scenario \pairedset{}.

\paragraph{Selection and rule review.}
Our construction log records the difficulty-based retention and the subsequent exclusions (\S\ref{sec:queries}). The input manifest contains the retained collection with its rule-derived gold labels. The completed scenario census flagged seven \mainset{} scenarios. An earlier inspection also flags \path{t3_concussed_teenager} for review of its early-violation premise. Dataset size counts both base templates and generated inputs; neither count substitutes for this rule review.

\section{Scoring, Agreement, and Reanalysis}
\label{app:scoring}
Final labels are reconstructed by joining the scenario, condition, response, and judgment tables in \path{review.db}. Label files are applied in this order: \path{labels_plan_sonnet_v2.jsonl}, \path{labels_luna_v2.jsonl}, and \path{corrections_v2.jsonl}. The duration answers use \path{stage_a_scores.jsonl}. Only the 13 evaluated models enter the analysis. A complete \pairedset{} tuple requires usable \condA{}, \condB{}, \condC{}, and non-violation control outputs.

\paragraph{Label counts.}
The \mainset{} has 41,367 labels on \condB{} and \condC{} responses, of which the rule screen assigns 13,223, and 233 are corrections. 
The 233 comprise the 173 main-set corrections produced by the adjudication pass of \S\ref{sec:evaluation}, which also produced the nine \pairedset{} corrections (182 in total), and 60 retained from two earlier single-judge precision audits (47 and 13); the correction files hold 368 rows in all, the remainder falling on models or inputs outside the evaluated sets.
The \pairedset{} has 1,284 labels on \condB{}, \condC{}, and non-violation control responses, of which the rule screen assigns 298, and nine are corrections. The reconstructed numerators and denominators match the recorded scores for all 13 models.

\paragraph{Uncertainty.}
Intervals resample whole templates with 10,000 bootstrap draws, keeping all inputs, conditions, and models of a template together. The seed is 20261005 for the \mainset{} and 20261004 for the \pairedset{}. The intervals measure variation across scenarios with the set of evaluated models held fixed.

\paragraph{Shared \pairedset{} coverage.}
The common 32-scenario subset contains 416 tuples: \condA{} is correct in 352, \condB{} catches 140, and \condC{} catches 266. \condB{} misses among correct \condA{} answers are 216/352 (61.4\%), preserving the \pairedset{} pattern after the incomplete scenario is removed.

\paragraph{Agreement on the \pairedset{}.}
The two judges overlap on 986 responses: 428 \condB{} violations, 428 \condC{} violations, and 130 non-violation controls. Table~\ref{tab:judges} reports their original four-category labels before the nine final corrections. The judges agree on 939/986 responses (95.23\%; Cohen's $\kappa=0.9047$). This overlap covers 76.8\% of the 1,284 labeled responses. The other 298 controls receive rule-screened \Proceed{} labels.

\begin{table}[t]
\centering
\resizebox{1.0\linewidth}{!}{
\begin{tabular}{lrrrr}
\toprule
 & \Catch{} & \Hedge{} & \Ask{} & \Proceed{} \\
\midrule
\Catch{} & 401 & 1 & 0 & 23 \\
\Hedge{} & 5 & 0 & 1 & 9 \\
\Ask{} & 0 & 0 & 0 & 0 \\
\Proceed{} & 5 & 2 & 1 & 538 \\
\bottomrule
\end{tabular}
}
\caption{Original judge labels. Rows: Claude Sonnet~5; columns: GPT-5.6 Luna.}
\label{tab:judges}
\end{table}

Collapsing the labels to \Catch{}/non-\Catch{} yields agreement of 952/986 (96.55\%) and $\kappa=0.9294$. Agreement measures consistency between automated judges. Further annotation must inspect responses on which the two judges agree and controls screened by rules.

\paragraph{Human audit.}
\label{app:human}
A pre-registered human-annotation study was designed and collected before this paper was drafted. The annotator worked blind in two phases, committing each label before the recorded label was revealed. Within the \pairedset{}, the sample covers 144 judged responses spanning all 33 scenarios and all 13 models, 21 adversarially flagged \Catch{} labels, six rule-screened controls, and 20 duration-parser checks. The study also includes a census of all 161 scenarios (the 128 \mainset{} templates and the 33 \pairedset{} scenarios), in which every one of the 33 \pairedset{} scenarios was kept with all six checklist items affirmed; the seven scenarios flagged by the annotator belong to the \mainset{}. The same audit reports $\kappa=0.53$ on the \mainset{} labels used in this paper.
Table~\ref{tab:human-audit} reports agreement with the recorded labels by
stratum.

Beyond label agreement, the annotator confirmed 19 of the 21 flagged \Catch{} labels (90.5\% [71.1, 97.4]), found one non-\Proceed{} reply among the six rule-screened controls, matched parser v7.1 on 17 of the 20 duration probes, and, on the two corrected responses in the sample, sided with the corrected label in the \Catch{} sense both times.

The non-violation controls are the audit's weakest stratum: the annotator read seven
judge-labeled control replies, and four rule-screened ones, as asserting
the removed mismatch where the recorded label says none. If those readings
survive adjudication, the reported false-alarm rate is a floor. They are the
first targets for the remaining annotation.

The audit provides single-annotator agreement with the recorded labels. The calibration consensus equaled the judge's own labels, so calibration demonstrates rubric fit rather than independent anchoring; and the reveal-stage agree-clicks, which enter no statistic, contradict the committed label on 75 of 360 draw items, bounding the weight the reveal stage can carry. Building a gold standard remains future work: blind double annotation of the \pairedset{} replies and stratified \mainset{} replies, scored for recognition, action adjustment, and usefulness, adjudicating the control replies the annotator read as false alarms.

\begin{table}[t]
\centering
\resizebox{1.0\linewidth}{!}{
\begin{tabular}{@{}lcccc@{}}
\toprule
& $\kappa$ & 95\% CI & $n$ & Agr.\ (\%) \\
\midrule
All sampled (4-label)      & 0.708 & 0.60--0.82      & 144 & 84.7 \\
Judge-labeled, vs.\ final  & 0.735 & 0.62--0.84      & 126 & 85.7 \\
\begin{tabular}[t]{l}Violations\\\quad(\condB{}\\\quad\&\condC{})\end{tabular} & 0.768 & 0.65--0.88      & 114 & 87.7 \\
\condC{} & 0.846 & 0.71--0.96      & 49  & 91.8 \\
\condB{} violations & 0.707 & 0.54--0.87      & 65  & 84.6 \\
Non-violation controls & 0.330 & $-$0.06--0.69   & 30  & 73.3 \\
\bottomrule
\end{tabular}
}
\caption{Human--LLM agreement with the recorded labels. $\kappa$ is Cohen's $\kappa$; Agr. is agreement.}
\label{tab:human-audit}
\end{table}

\end{document}